\documentclass{article}

\usepackage{PRIMEarxiv}

\usepackage[utf8]{inputenc} 
\usepackage[T1]{fontenc}    
\usepackage{hyperref}       
\hypersetup{
    colorlinks=true,
    linkcolor=blue,
    filecolor=cyan,      
    urlcolor=magenta,
    }
\usepackage{url}            
\usepackage{booktabs}       
\usepackage{amsfonts,amssymb,amsmath}       
\usepackage{nicefrac}       
\usepackage{microtype}      
\usepackage{lipsum}
\usepackage{fancyhdr}       
\usepackage{graphicx}       
\usepackage{float}
\usepackage{wrapfig}
\usepackage{xcolor}
\usepackage{soul}
\usepackage[nameinlink]{cleveref}
\graphicspath{{Figures/}}     
\usepackage{subcaption}
\usepackage{caption}
\usepackage{vcell}
\usepackage{colortbl}
\usepackage{makecell}

\usepackage{romannum}
\AtBeginDocument{\pagenumbering{arabic}}

\usepackage{framed,multirow}

\usepackage{pifont}
\usepackage{amsfonts,amssymb,amsmath}       
\usepackage{latexsym}
\usepackage{comment}

\usepackage{tabularray}

\UseTblrLibrary{booktabs}

\definecolor{SandyBeach}{rgb}{1,0.917,0.776}
\definecolor{NavajoWhite}{rgb}{1,0.882,0.666}
\definecolor{Tradewind}{rgb}{0.36,0.678,0.678}
\definecolor{DeYork}{rgb}{0.541,0.764,0.541}
\definecolor{ColonialWhite}{rgb}{1,0.909,0.752}
\definecolor{Grandis}{rgb}{1,0.843,0.564}
\definecolor{GoldenTainoi}{rgb}{1,0.772,0.36}
\definecolor{YellowSea}{rgb}{1,0.666,0.05}
\definecolor{WebOrange}{rgb}{1,0.647,0}
\definecolor{PinkLace}{rgb}{1,0.905,1}
\definecolor{PinkLace1}{rgb}{1,0.788,1}
\definecolor{llightgreen}{RGB}{225,250,225}
\definecolor{llightred}{RGB}{255,215,215}
\definecolor{llightcyan}{RGB}{215,255,250}
\definecolor{LavenderRose}{rgb}{1,0.607,1}
\definecolor{lgray}{rgb}{0.88,0.88,.88}
		
\definecolor{lightpurple}{RGB}{213,172,213}
\definecolor{llightpurple}{RGB}{237,221,237}
		
\definecolor{lightpink}{RGB}{255,205,213}
\definecolor{llightpink}{RGB}{255,234,237}							

\definecolor{lyellow}{RGB}{255,232,141}
\definecolor{llyellow}{RGB}{255,244,197}	

\definecolor{skyblue}{RGB}{135,206,235}
\definecolor{lskyblue}{RGB}{215,235,255}
\AtBeginDocument{\pagenumbering{arabic}}

\title{Foundational Models in Medical Imaging: A Comprehensive Survey and Future Vision}

\vspace{-5em}
\author{
	Bobby Azad \\
	Electrical Engineering and Computer Science Department\\
	South Dakota State University \\
	Brookings, USA \\ 
    \And
	Reza Azad \\
	Faculty of Electrical Engineering and Information Technology\\ 
	RWTH Aachen University \\
	Aachen, Germany \\ 
    \And
	Sania Eskandari \\
	Department of Electrical Engineering \\
	University of Kentucky \\
	Lexington, USA \\ 
    \And
	Afshin Bozorgpour \\
	Faculty of Informatics and Data Science\\
	University of Regensburg \\
	Regensburg, Germany \\ 
    \And
    Amirhossein Kazerouni \\
	School of Electrical Engineering \\
	Iran University of Science and Technology \\
	Tehran, Iran\\ 
    \And
 	Islem Rekik \\
	BASIRA Lab, Imperial-X and Computing Department \\ 
	Imperial College London \\
	London, UK \\ 
    \And
    Dorit Merhof \thanks{Corresponding author: Dorit Merhof,  dorit.merhof\@ur.de} \\
    Faculty of Informatics and Data Science \\ 
	University of Regensburg \\
	Regensburg, Germany
}

\begin{document}
\maketitle

\vspace{-2em}
\begin{abstract}
Foundation models, large-scale, pre-trained deep-learning models adapted to a wide range of downstream tasks have gained significant interest lately in various deep-learning problems undergoing a paradigm shift with the rise of these models. Trained on large-scale dataset to bridge the gap between different modalities, foundation models facilitate contextual reasoning, generalization, and prompt capabilities at test time. The predictions of these models can be adjusted for new tasks by augmenting the model input with task-specific hints called prompts without requiring extensive labeled data and retraining.
Capitalizing on the advances in computer vision, medical imaging has also marked a growing interest in these models. With the aim of assisting researchers in navigating this direction, this survey intends to provide a comprehensive overview of foundation models in the domain of medical imaging. 
Specifically, we initiate our exploration by providing an exposition of the fundamental concepts forming the basis of foundation models. Subsequently, we offer a methodical taxonomy of foundation models within the medical domain, proposing a classification system primarily structured around training strategies, while also incorporating additional facets such as application domains, imaging modalities, specific organs of interest, and the algorithms integral to these models. Furthermore, we emphasize the practical use case of some selected approaches and then discuss the opportunities, applications, and future directions of these large-scale pre-trained models, for analyzing medical images. In the same vein, we address the prevailing challenges and research pathways associated with foundational models in medical imaging. These encompass the areas of interpretability, data management, computational requirements, and the nuanced issue of contextual comprehension. Finally, we gather the over-viewed studies with their available open-source implementations at our \href{https://github.com/mindflow-institue/Awesome-Foundation-Models-in-Medical-Imaging}{GitHub}. We aim to update the relevant latest papers within it regularly.
\end{abstract}

\keywords{Foundation models \and Deep learning \and  Language and vision \and Large language models \and Score-based models \and Self-supervised learning \and Medical applications \and Survey}

\section{Introduction}
\label{intro}
Medical imaging is at the forefront of healthcare, playing a pivotal role in diagnosing and treating diseases. Recent advancements in Artificial Intelligence (AI) have given rise to a new era in the field of medical imaging, driven by the development of Foundation Models (FMs). 
Foundation Models (FMs) are a type of artificial intelligence (AI) model that exhibit significant progress in their development. These models are typically trained on extensive, diverse dataset, frequently utilizing self-supervision techniques on a massive scale. Following this initial training, they can be further adapted, such as through fine-tuning, for a wide array of downstream tasks that are related to the original training data \cite{bommasani2021opportunities}.

\begin{figure}[!ht]
	\centering
	\includegraphics[width=.45\columnwidth]{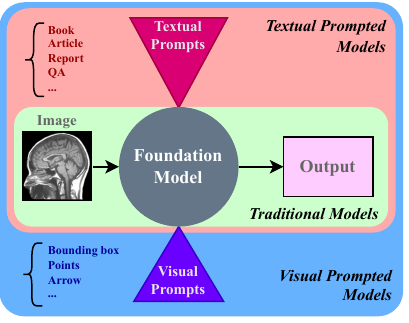}
    \caption{Visual illustration of how our extensive classification categorizes existing works into textually and visually prompted models, distinct from traditional vision models.}
    \label{fig:models}
\end{figure}

In contrast to the conventional deep learning paradigm, which heavily relies on large-scale, task-specific, and crowd-labeled data to train individual deep neural networks (DNNs) for various visual recognition tasks, FMs provide a more efficient alternative. They are pretrained on large-scale dataset that are nearly unlimited in availability, enabling straightforward application to downstream tasks with only a limited amount of labeled data. This shift in approach shows potential for significantly decreasing the labor and time usually necessary for such tasks. 
The recent surge can be attributed to the progress made possible by large language models (LLMs) and the expansion of data and size \cite{awais2023foundational}. Models such as GPT-3 \cite{brown2020language}, PaLM \cite{chowdhery2022palm}, Galactica \cite{taylor2022galactica}, and LLaMA \cite{touvron2023llama} have exhibited strong ability to comprehend natural language and solve complex tasks with zero/few-shot learning, attaining remarkable results without requiring extensive task-specific data.
Large-scale vision foundation models are currently making significant advances in perception tasks, as highlighted by~\cite{jia2021scaling,yao2021filip}. Specifically, vision-language models (VLM) are pre-trained with large-scale image-text pairs and are then directly applicable to downstream visual recognition tasks. VLMs generally consist of three fundamental parts: textual features, visual features, and a fusion module. These elements work together in harmony, allowing the models to efficiently use text and visual data to generate contextually appropriate and logical results. Specifically, the pre-training of VLMs typically adheres to vision-language objectives that aid in acquiring image-text correlations from large collections of image-text pairs. For instance, the pioneer study CLIP \cite{schuhmann2021laion}, an image-text matching
model, utilizes contrastive learning methods to generate fused representations for images and texts. The learning objective is to minimize the gap between the representation of an image and its corresponding text, while simultaneously increasing the separation between the representations of unrelated pairs. In addition to this so-called ``\emph{\textbf{Textually Prompted Models (TPMs)}}", researchers have also explored Feature Maps (FMs) that can be prompted by visual inputs (points, boxes, masks) which we refer to as ``\emph{\textbf{Visually Prompted Models (VPMs)}}" \cite{awais2023foundational} (see \cref{fig:models} for a visual depiction of both). Recently, the Segment Anything (SA) model \cite{kirillov2023segment} has garnered significant attention in the vision community. SAM is a promptable model developed for the purpose of broad image segmentation. It was trained using a promptable segmentation task that enables powerful zero-shot generalization on 11 million images and over 1 billion masks. Furthermore, SAM has been expanded and refined through training on a large dataset, which encompasses 4.6 million medical images and 19.7 million corresponding masks \cite{cheng2023sammed2d}. This dataset offers a rich diversity, covering 10 distinct medical data modalities, and featuring annotations for 4 anatomical structures in addition to lesions. The training regimen is comprehensive, representing 31 major human organs. Notably, it has yielded impressive results that have bolstered the model's capacity for enhanced generalization.

Moreover, this generic visual prompt-based segmentation model has recently been adapted to a wide range of downstream tasks, including medical image analysis \cite{ma2023segment,lei2023medlsam}, image inpainting \cite{yu2023inpaint}, style transfer \cite{liu2023any}, and image captioning \cite{wang2023caption} to name a few. Apart from foundational models that rely on textual and visual prompts, research endeavors have also delved into creating models that harmonize various types of paired modalities (such as image-text, and video-audio) to learn representations assisting diverse downstream tasks.

\begin{figure*}[!thb]
\centering
    \begin{subfigure}[b]{0.33\textwidth}
         \centering
         \includegraphics[width=\textwidth]{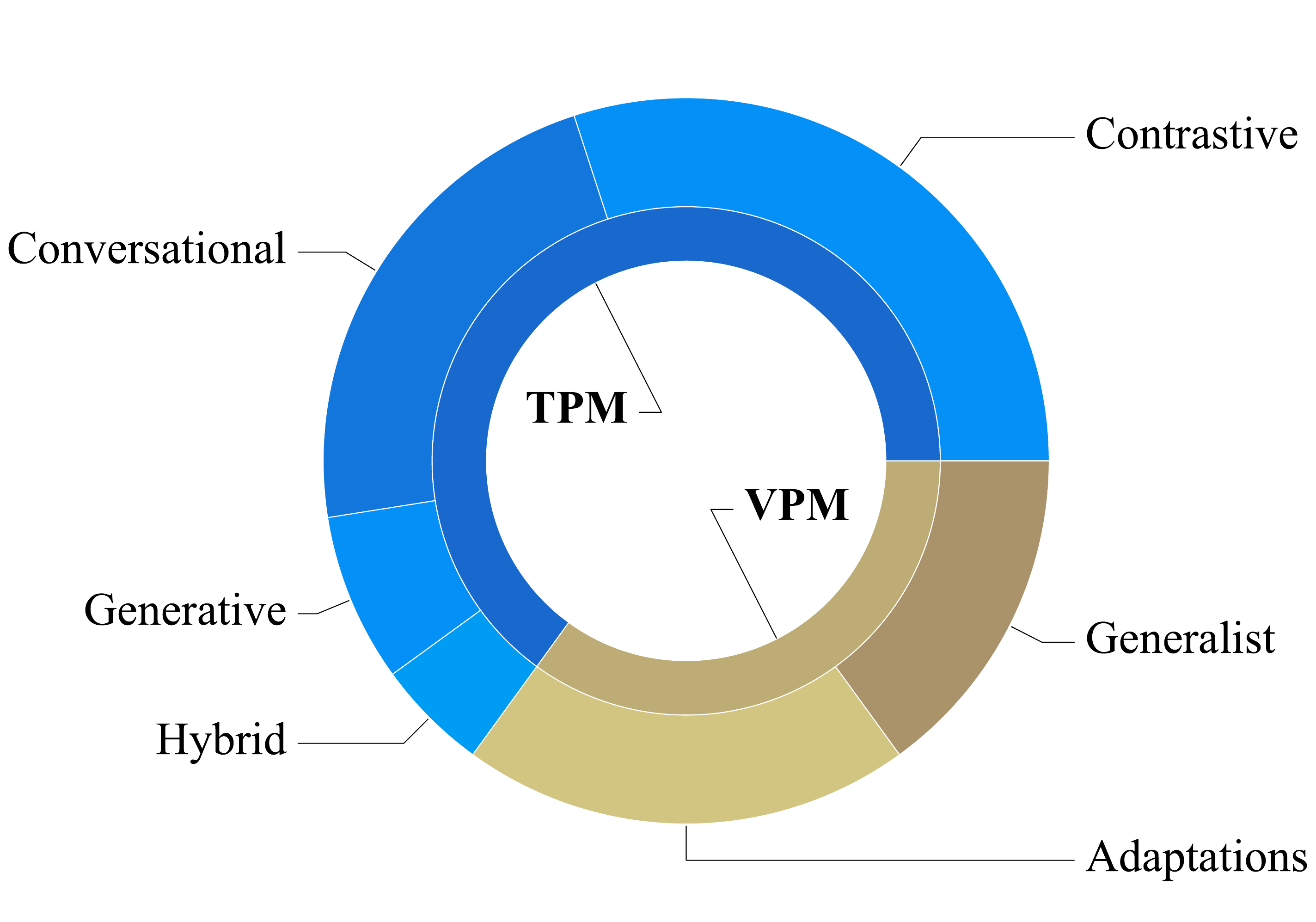}
         \caption{Algorithms}
     \end{subfigure}
     \hfill
     \begin{subfigure}[b]{0.33\textwidth}
         \centering
         \includegraphics[width=\textwidth]{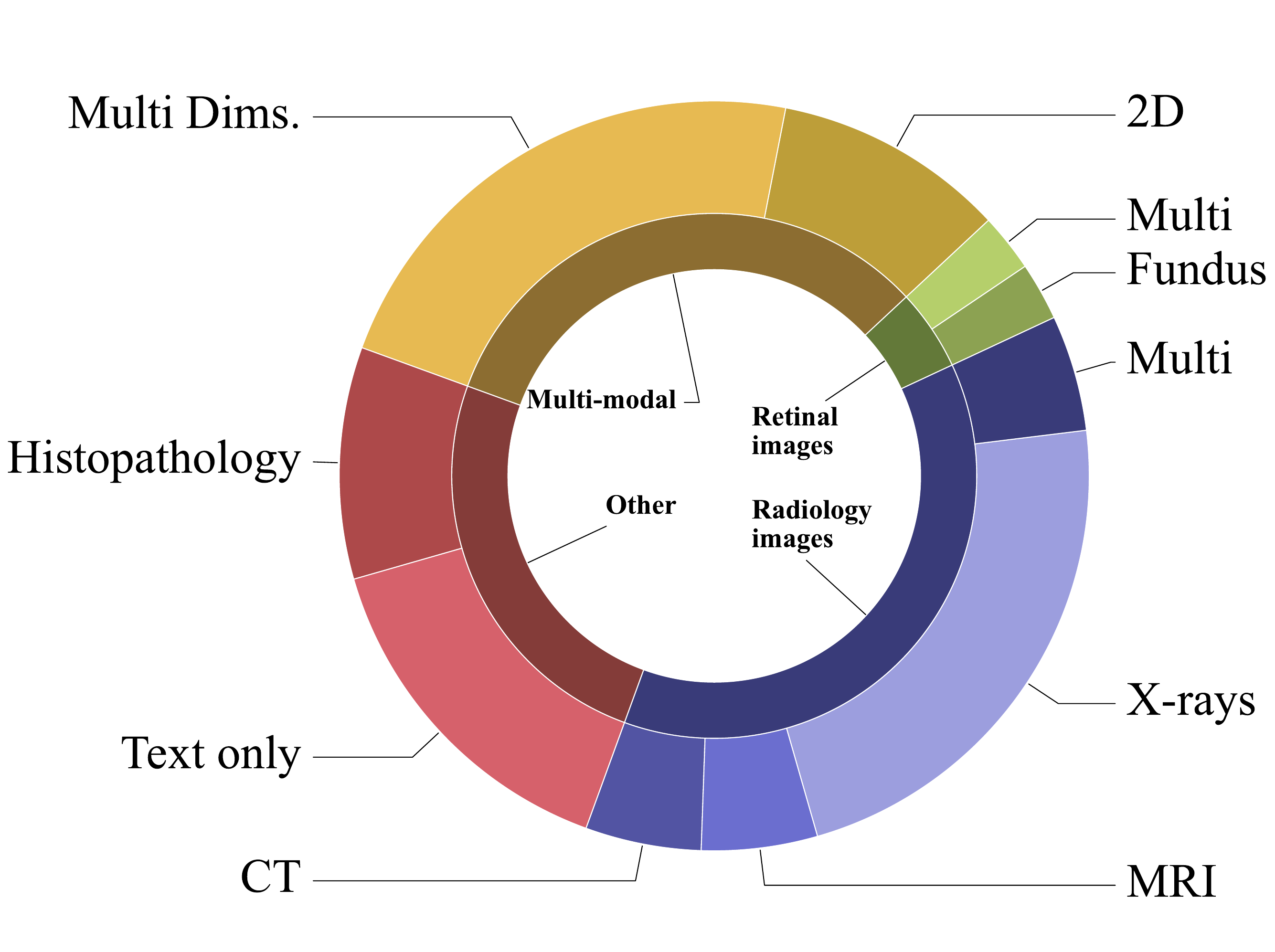}
         \caption{Modalities}
     \end{subfigure}
     \hfill
     \begin{subfigure}[b]{0.33\textwidth}
         \centering
         \includegraphics[width=\textwidth]{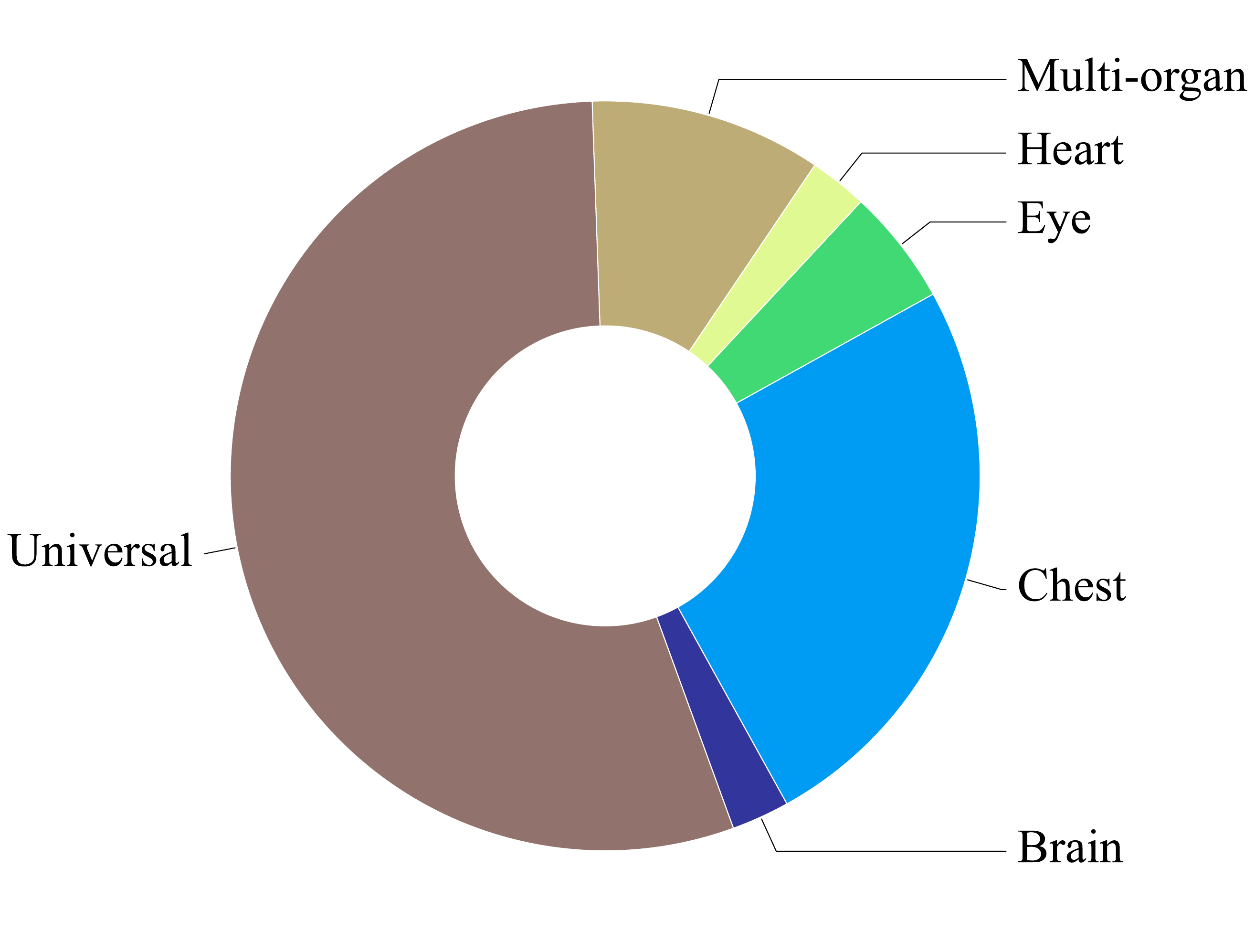}
         \caption{Organs}
     \end{subfigure}
\caption{The diagram (a) displays the distribution of published papers categorized by their algorithm, (b) categorizes them by their imaging modalities, and (c) classifies them by the type of organ concerned. It is worth noting that the total number of papers included in the analysis is 40.}
\label{fig:graphs}
\end{figure*}

The creationn of foundation models has garnered significant attention in the medical AI system development realm \cite{ji2023continual,zhang2023text,azad2023advances,wang2023foundation,nguyen2023lvm}. Despite the substantial advancements in biomedical AI, the main methodologies used still tend to be tas-specific models. However, medical practice encompasses various data modalities comprising text, imaging, genomics, and others, making it essentially multimodal \cite{tu2023generalist}. Inherently, a \emph{Medical Foundation Model (MFM)} has the ability to adaptively interpret various medical modalities, including diverse data sources such as images, electronic medical records, lab findings, genomic information, medical diagrams, and textual data \cite{moor2023foundation}.
Hence, foundational models have the potential to provide an enhanced foundation for addressing clinical issues, advancing the field of medical imaging, and improving the efficiency and effectiveness of diagnosing and treating diseases, leading to the opportunity to develop a unified biomedical AI system that can interpret complex multimodal data. Due to the acceleration of both biomedical data production and advancements, the influence of these models is expected to expand due to an influx of contributions. As shown in~\Cref{fig:graphs}, a significant body of research has been devoted to the application of FMs in diverse medical imaging contexts until the first release of our survey in October 2023. These contributions encompass a wide range of potential applications, from fundamental biomedical discoveries to the upgrading of healthcare delivery. Hence, it is advantageous for the community and timely to review the existing literature. 

This paper provides a holistic overview of the foundation models developed for medical imaging applications. We distinguish existing works inspired by the taxonomy proposed in \cite{awais2023foundational} and highlight the major strengths and shortcomings of the existing methods. We hope that this work will
point the way forward, provide a roadmap for researchers, stimulate further interest and enthusiasm within the vision community, and harness the potential of foundation models in the medical discipline. This survey will be regularly updated to reflect the dynamic progress of the MFMs, as
this is a rapidly evolving and promising field towards AGI in the biomedical field. Our major contributions include:

$\bullet$~We conduct a thorough and exhaustive examination of foundation models proposed in the field of medical imaging, beginning from background and preliminaries for foundation models, to specific applications along with the
organ concerned and imaging modality in a hierarchical and structured manner

\begin{figure*}[!htb]
\centering
\includegraphics[width=\textwidth]{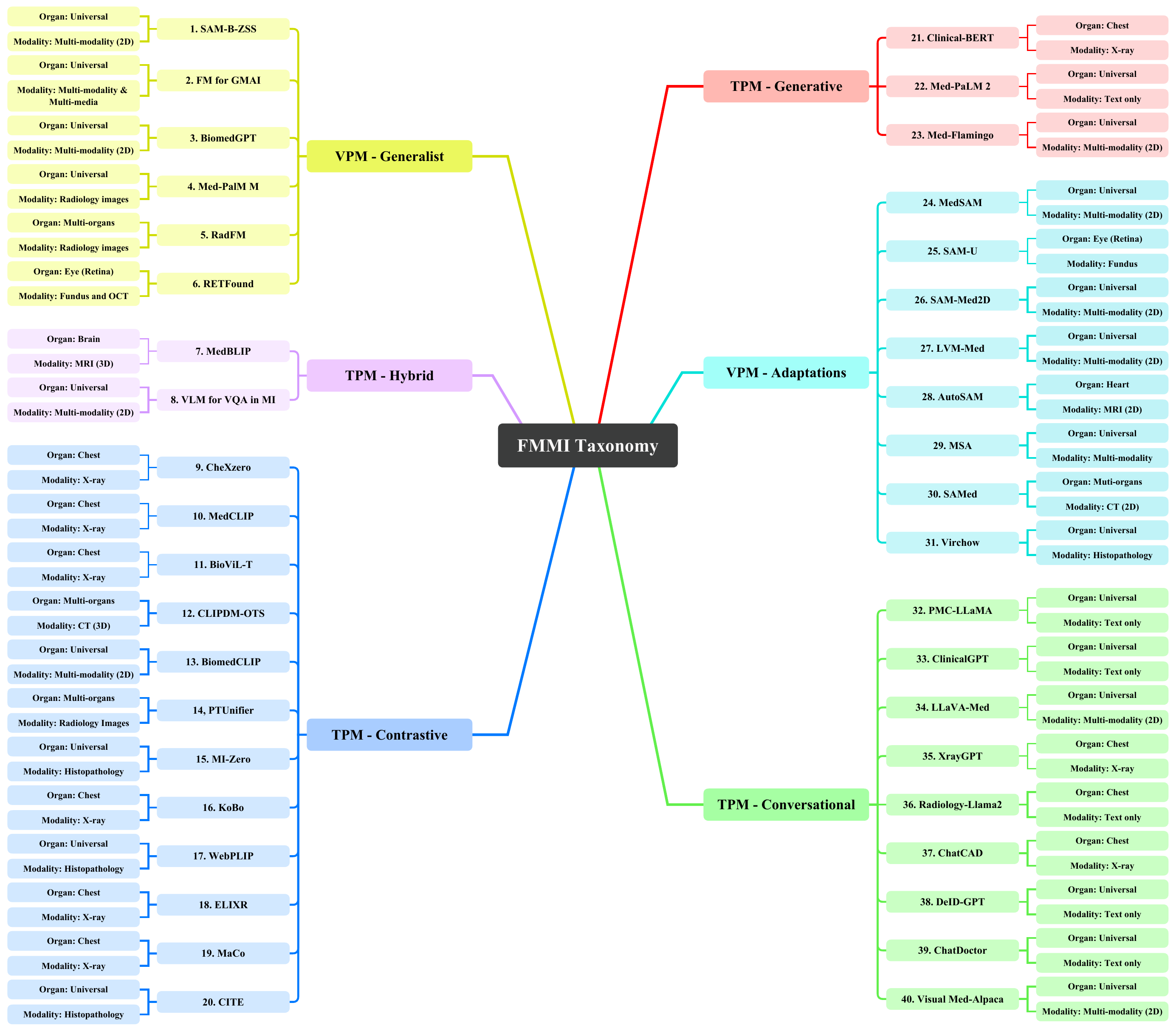}
\caption{The suggested taxonomy for foundational models used in medical imaging research consists of six distinct groups: I) VPM-Generalist, II) TPM-Hybrid, III) TPM-Contrastive, IV) TPM-Generative, V) VPM-Adaptations, and VI) TPM-Conversational. To maintain conciseness, we assign ascending prefix numbers to each category in the paper's name and cite each study accordingly as follows: \protect\input{taxonomy-references} }
\label{fig:taxonomy}
\end{figure*}

$\bullet$~Our work provides a taxonomized (\Cref{fig:taxonomy}), in-depth analysis (e.g. task/organ-specific research progress and limitations), as well as a discussion of various aspects.

$\bullet$~Furthermore, we discuss the challenges and unresolved aspects linked to foundation models in medical imaging. We pinpoint new trends, raise important questions, and propose future directions for further exploration.

\subsection{Clinical Importance}
In medical imaging, foundation models are reshaping the way research methods are designed and paradigms are approached, paving the way for innovative advancements and pioneering breakthroughs across various sectors owing to some of their inherent properties aligned with the medical domain as follows. \\
\textbf{Multi-Modality:} Despite advances in biomedical AI, most models today are limited to single-task, unimodal functions. For instance, a mammogram interpretation AI excels at breast cancer screening but can't incorporate patient records, and additional data like MRI, or engage in meaningful dialogue, limiting its real-world applicability. \\
\textbf{Explainability and Generalization:} The absence of explainability in deep learning models can erode trust among clinicians accustomed to clear clinical insights \cite{kazerouni2023diffusion}. The ability of models to generalize across different medical settings is vital due to varying data sources. Foundation models address these issues by offering a unified framework for tasks like detection and classification, often trained on diverse datasets from various medical centers, enhancing their potential for clinical use by ensuring interpretability and broad applicability. \\
\textbf{Privacy Preservation:} The computer vision community has a history of open-sourcing datasets, but in medical imaging, privacy regulations limit data sharing. Foundation models offer a privacy-preserving alternative by allowing knowledge transfer without direct access to sensitive data. Additionally, federated learning enables model training on distributed data while keeping it on local machines, ensuring data privacy. Moreover, foundation models facilitate privacy preservation by generating synthetic data resembling real medical images, eliminating the need for actual patient data in model training. \\
\textbf{Adaptability:} Existing medical AI models struggle when faced with distribution shifts caused by changes in technology, procedures, settings, or populations. In contrast, MFMs can effectively adapt to these shifts through in-context learning. For instance, a hospital can teach an MFM model to interpret X-rays from a new scanner by providing a few examples as prompts, enabling it to adjust to new data distributions in real time. This capability is mainly seen in large language models and is not common in conventional medical AI models, which would typically require complete retraining with new datasets. \\
\textbf{Domain Knowledge:} Unlike clinicians, traditional medical AI models often lack initial medical domain knowledge, relying solely on statistical associations. Medical imaging foundation models like GMAI can address this limitation by integrating formal medical knowledge, using structures like knowledge graphs and retrieving relevant context from existing databases, improving their performance on specific tasks.
In summary, foundation models play a crucial role in advancing medical applications by providing a robust and adaptable framework that enhances efficiency, generalizability, and privacy preservation. Their ability to support various clinical tasks and promote collaboration makes them invaluable tools for improving patient care and medical research.

\subsection{Relevant Surveys}
With the recent success of foundation models, there has
been a surge of surveys and contributions in this domain.
Some of the reviews investigate recent advances
in LLMs, distinguishing different aspects of LLMs by analyzing the impact of pre-training adaptation tuning, utilization, and evaluation \cite{zhao2023survey,fan2023bibliometric,huang2022towards}. In the context of vision models, the work of \cite{awais2023foundational} provides a comprehensive review of FMs including their typical architecture design, training objectives, and prompting mechanisms. The work of \cite{gu2023systematic} delivers a comprehensive survey of research in prompt engineering on diverse types of vision-language models, organizing existing prompt-engineering approaches from a new perspective. Besides, \cite{zhang2023vision} provides a systematic review of visual language models for various visual recognition tasks including image classification, object detection, and semantic segmentation.
In the medical imaging field, \cite{moor2023foundation} identifies the potential applications, opportunities, and challenges of MFMs. The work of \cite{shi2023generalist} provides a comprehensive and objective evaluation of SAM on medical image segmentation, while \cite{zhang2023challenges} discusses the spectrum, and future directions of foundation models.
However, different from the aforementioned works, we devise a multi-perspective taxonomy of foundation models in the medical community, providing a systematical category of research in medical foundation models and their applications dividing them into textually prompted models, and visually prompted models where each paper is broadly classified according to the proposed algorithm along with the organ concerned and imaging modality, respectively. We present the concepts and theoretical foundations behind foundation models ranging from training objectives and instruction-aligning to prompt engineering (\Cref{sec:background}). In \Cref{sec:textual}, we comprehensively cover an extensive and up-to-date overview of the recent medical foundation models, as shown in \Cref{fig:taxonomy}. We wrap up this survey by pinpointing future directions and open challenges facing foundation models in medical imaging in \Cref{sec:challenge}.

\subsection{Search Strategy}
We conducted extensive searches across various platforms, such as DBLP, Google Scholar, and Arxiv Sanity Preserver. We leveraged their search capabilities to create tailored queries and compile comprehensive lists of academic works. These searches encompassed a broad spectrum of scholarly publications, including peer-reviewed journal articles, conference papers, workshop materials, non-peer-reviewed content, and preprints. We tailored our search criteria to achieve this diversity. Our specific search queries consisted of keywords \texttt{(foundation* $|$ generalist* $|$ medical* $|$ \textbf{\{Task\}}*), (med-\textbf{\{FM\}} $|$ foundation*), (foundation* $|$ biomedical* $|$ image* $|$ model*)}, where \textbf{\{FM\}} and \textbf{\{Task\}} refer to one well-known vision foundation model(such as \textit{PaLM}, \textit{CLIP}, etc) or Tasks (such as \textit{Segmentation}, \textit{Question Answering}, etc) in medical imaging. We then applied filtering to eliminate false positives, ensuring that only papers related to foundation models were included in our analysis.

\subsection{Paper Organization.}
The rest of the survey is organized as follows. \Cref{sec:background} presents the background and preliminaries for foundation models. We adopt the taxonomy of \cite{awais2023foundational} and categorize previous studies into two main groups: those prompted by textual inputs (discussed in \cref{sec:textual}) and those driven by visual cues (discussed in \cref{sec:visual}). In the context of textually prompted foundation models, we further subdivide them into contrastive, generative, hybrid (combining contrastive and generative approaches), and conversational visual language models. In addition, we differentiate textually prompted models into adaptations and generalist models. Furthermore, \Cref{sec:challenge} reveals the risk, open problems, and
future directions of foundation models. Finally, we conclude our research in \Cref{sec:conclusion}.

\section{Preliminaries}
\label{sec:background}

The term "foundational models" made its debut at Stanford Institute for Human-Centred AI in \cite{bommasani2021opportunities} with the definition of \emph{"the
base models trained on large-scale data in a self-supervised or
semi-supervised manner that can be adapted for several other
downstream tasks"}. Specifically, inspired by the surge of large language models (LLMs), using the basic fundamentals of deep learning such as DNNs and self-supervised learning, foundation models have emerged by massively scaling up both data and model size. 
In this section, we introduce the basic model architectures, concepts, and settings behind FMs focusing on contributing factors for these models in computer vision such as training
objectives, instruction-aligning, inference procedure and prompting.

\subsection{Pre-training Objectives}
\label{sec:objective}
Diverse pretraining objectives have been devised to learn a rich understanding of the relationship between vision and language \cite{mu2021slip,9878693,he2020momentum,radford2021learning}. We broadly categorize them into contrastive and generative objectives.

\subsubsection{Contrastive Objectives}

Contrastive objectives instruct models to acquire distinctive representations \cite{chen2020simple,he2020momentum} by bringing related sample pairs closer together while pushing unrelated pairs farther apart within the feature space. Specifically, \textbf{\textit{Image Contrastive Loss (ICL)}} aims to learn discriminative image features making a query image closely resemble its positive keys (i.e., its data augmentations) while ensuring it remains distant from its negative keys (i.e., other images) within the embedding space. Consider a batch of $B$ images, contrastive objectives such as InfoNCE \cite{Oord2018RepresentationLW} and its variations \cite{he2020momentum,chen2020simple}, $\mathcal{L}_{\text{I}}^{\text{InfoNCE}}$ can be expressed as:

\[
\mathcal{L}_{\text{I}}^{\text{InfoNCE}} = -\frac{1}{B} \sum_{i=1}^B \log \frac{\exp\left(\theta_i^{\text{query}} \cdot \theta_{+}^{\text{positive}} / \tau\right)}{\sum_{j=1, j \neq i}^{B+1} \exp\left(\theta_i^{\text{query}} \cdot \theta_j^{\text{key}} / \tau\right)}
\]

\noindent where $\theta_i^{\text{query}}$ represents the query embedding, $\{\theta_j^{\text{key}}\}_{j=1, j \neq i}^{B+1}$ are the key embeddings, where $\theta_{+}^{\text{positive}}$ denotes the positive key corresponding to $\theta_i^{\text{query}}$, while the rest are considered negative keys. The hyperparameter $\tau$ governs the density of the learned representation.

\textbf{\textit{Image-Text Contrastive Loss (ITCL)}}
Seeks to develop distinctive image-text representations by bringing together the embeddings of matched images and texts and pushing apart those that do not match \cite{radford2021learning,jia2021scaling}.
Let $\left(i, t_i\right)$ represent the $i$-th image-text example, then, the image-to-text loss is calculated as:

\[
\mathcal{L}_{I \rightarrow T} = -\log \left[\frac{\exp \left(\theta_i \cdot \theta_{+} / \tau\right)}{\sum_{j=1}^N \exp \left(\theta_i \cdot \theta_j / \tau\right)}\right]
\]

\noindent where $N$ is the total number of such pairs, and $\theta_i$ corresponds to the embedding for image $i$, while $\theta_{+}$ and $\theta_j$ denote positive and negative text representations, respectively. The losses are computed with a focus on the relationship between images and texts while considering the temperature parameter $\tau$.

The text-to-image loss is also calculated similarly, and the total loss is the sum of these two terms:

\[
\mathcal{L}_{ITC} = \frac{1}{N} \sum_{i=1}^N \left[\mathcal{L}_{I \rightarrow T} + \mathcal{L}_{T \rightarrow I}\right]
\]

Akin to \textit{ICL} and \textit{ITCL}, various contrastive loss functions have also found an application (\textbf{SimCLR} \cite{chen2020simple,DBLP:conf/nips/ChenKSNH20}, \textbf{FILIP Loss} \cite{Yao2021FILIPFI}, \textbf{Region-Word Alignment (RWA)} \cite{li2022grounded}, and \textbf{Region-Word Contrastive (RWC)} \cite{zhang2022glipv2}).

\subsubsection{Generative Objectives}
Generative objectives involve teaching networks to produce image or text data, which allows them to acquire semantic features, accomplished through tasks like image generation \cite{Bao2021BEiTBP}, and language generation \cite{Liu2019RoBERTaAR}.

\textbf{\textit{Masked Image Modelling (MIM)}} 
involves the acquisition of cross-patch correlations by applying masking and image reconstruction techniques. In MIM, a selection of patches within an input image is randomly masked, and the encoder is trained to reconstruct these masked patches based on the unmasked patches. For a given batch of $B$ images, the loss function is formulated as:

\[
\mathcal{L}_{MIM} = -\frac{1}{B} \sum_{i=1}^B \log f_\theta\left(\bar{x}_i^I \,|\, \hat{x}_i^I\right),
\]

\noindent where $\bar{x}_i^I$ and $\hat{x}_i^I$ represent the masked and unmasked patches within $x_i^I$, respectively \cite{zhang2023vision}.

\textbf{\textit{Masked Language Modelling (MLM)}} is a widely adopted pretraining objective in Natural Language Processing (NLP). In MLM, a specific percentage of input text tokens is randomly masked, and these masked tokens are reconstructed using the unmasked ones. The loss function for MLM can be expressed as:

\[
\mathcal{L}_{MLM} = -\frac{1}{B} \sum_{i=1}^B \log f_\theta\left(\bar{x}_i^T \,|\, \hat{x}_i^T\right),
\]

\noindent where $\bar{x}_i^T$ and $\hat{x}_i^T$ denote the masked and unmasked tokens within $x_i^T$, respectively, and $B$ denotes the batch size \cite{zhang2023vision}.

Likewise, diverse additional generative loss functions have been introduced in the field including Masked Multimodal Modeling (MMM) loss \cite{singh2022flava}, Image-conditioned Masked Language Modeling (IMLM) loss \cite{zhang2023toward}, and Captioning with Parallel Prediction (CapPa) \cite{tschannen2023image}.

\subsection{Pre-training Tasks}

As discussed in \cref{sec:objective}, FMs pre-training has been studied with typical approaches including contrastive objectives, and generative objectives. In natural language processing, certain pre-training tasks include masked language modeling, where words in the input sequence are randomly hidden, and the model predicts these hidden words during pre-training. Another task involves next-sentence-prediction, where pairs of sentences from distinct documents are presented, and the model determines whether the order of these sentences is accurate. Additionally, there's the denoising auto-encoder task, which introduces noise into the original text corpus and then aims to reconstruct the pristine input using the noisy version of the corpus.
Likewise, to enable the generalization of learned representations to a range of downstream vision domains, pretext tasks such as inpainting \cite{wang2023images}, auxiliary supervised discriminative tasks, and data reconstruction tasks \cite{luo2023segclip} are used in the pre-training stage.

\subsection{Instruction-Aligning}

Instruction-aligning methods aim to let the LM follow human intents and generate meaningful outputs. This process involves either fine-tuning the model on a diverse set of tasks with human-annotated prompts and feedback (RLHF) \cite{ouyang2022training}, conducting supervised fine-tuning on publicly available benchmarks and datasets, which are augmented with manually or automatically generated instructions, and improving the reasoning ability
of LLMs by instructing them to produce a sequence of intermediate actions that ultimately lead to the solution of a multi-step problem (Chain-of-thought) \cite{bai2022constitutional}.

\subsection{Prompt Engineering}

Prompt engineering refers to a method that enhances a large pre-trained model by incorporating task-specific hints, referred to as prompts, to tailor the model for new tasks 
enabling the power to acquire
predictions based only on prompts without updating model parameters \cite{gu2023systematic}.
In the context of large language models (LLMs) prompting techniques can be categorized into two primary groups depending on the clarity of the templates they employ: "soft prompts" (optimizable, learnable) and "hard prompts (manually crafted text prompts)". Within the "hard prompt" category, there are four subcategories: task instructions, in-context learning, retrieval-based prompting, and chain-of-thought prompting. In contrast, "soft prompts" fall into two strategies: prompt tuning and prefix token tuning, which differ in whether they introduce new tokens into the model's architecture or simply attach them to the input. 
In the vision domain, prompt engineering facilitates the acquisition of joint multi-modal representations (e.g., CLIP \cite{radford2021learning} for image classification of ALIGN \cite{Li2021AlignBF}) to introduce human interaction to the foundational models and employs vision-language models for visual tasks.

\section{Foundational Models for Medical Imaging}
\label{chapter3}
Establishing a taxonomy for foundational models in medical imaging analysis follows the standard practices commonly employed in the field. However, we distinguish our approach by providing extensive additional information for each sub-category as presented in \Cref{fig:taxonomy}. In this section, we explore foundational-based methods, which have been introduced to tackle diverse challenges in medical imaging analysis through the design of distinct training strategies.

\subsection{Textually Prompted Models}
\label{sec:textual}

\subsubsection{Contrastive}
Contrastive textually prompted models are increasingly recognized in the foundational models for medical imaging. They learn representations that encapsulate the semantics and relationships between medical images and their textual prompts. Leveraging contrastive learning objectives, they draw similar image-text pairs closer in the feature space while pushing dissimilar pairs apart. These models are pivotal for image classification, segmentation, and retrieval tasks. Architectural explorations have ranged from dual-encoder designs—with separate visual and language encoders—to fusion designs that merge image and text representations via decoder and transformer-based architectures. Their potential in medical imaging tasks such as lesion detection, disease classification, and image synthesis is evident in numerous studies. In this direction, Wang et al. \cite{wang2022medclip} introduced the MedCLIP framework, demonstrating its superiority over state-of-the-art methods in zero-shot prediction, supervised classification, and image-text retrieval. Expanding upon the success of models like CLIP, \cite{zhang2023large} unveiled BiomedCLIP, tailored for biomedical vision-language processing. Its training on a vast dataset of 15 million figure-caption pairs highlighted the efficacy of specialized pretraining in the medical imaging field.

\begin{figure*}[!thb]
\centering
\includegraphics[width=0.95\textwidth]{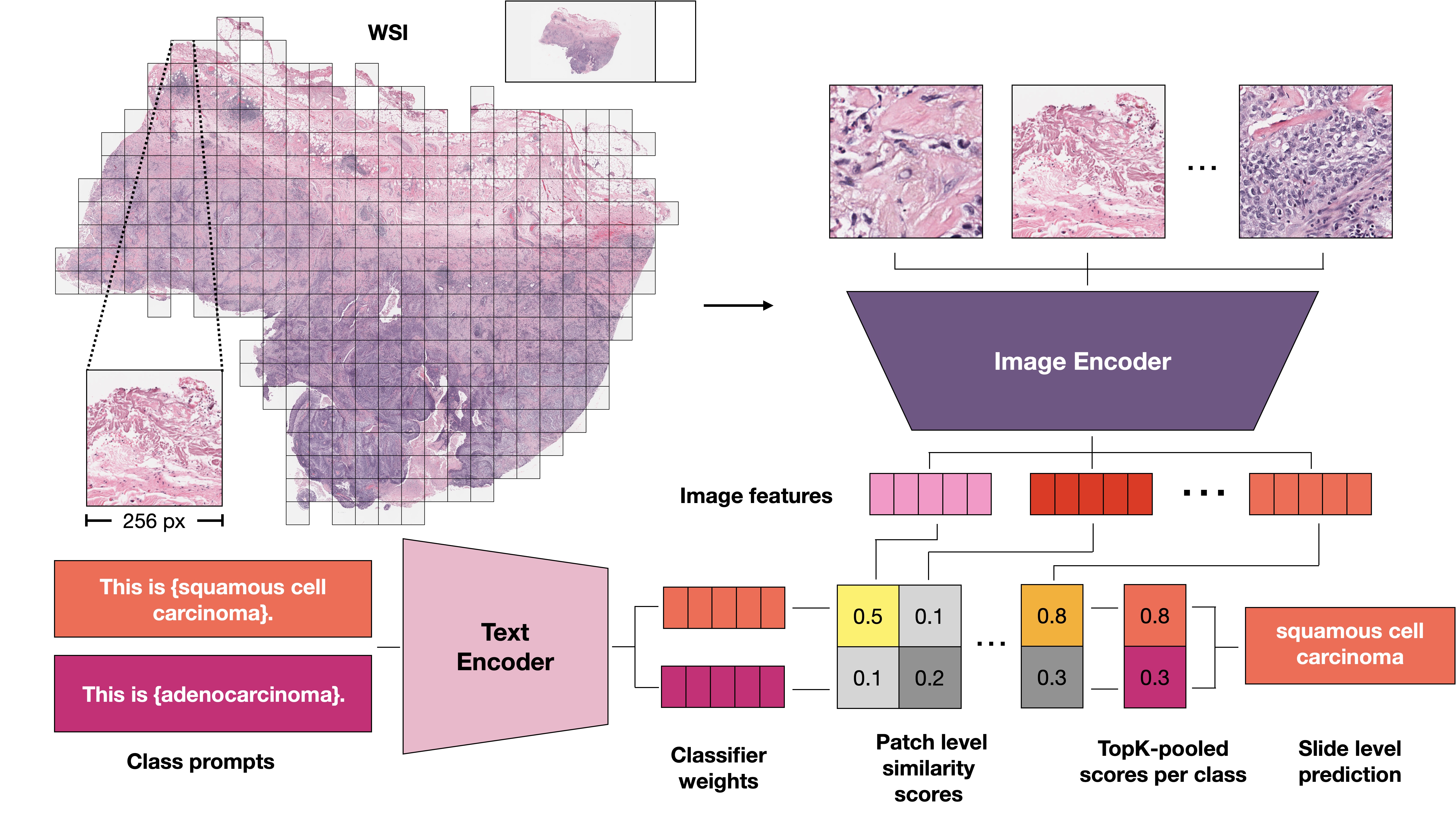}
\caption{Schematic of MI-Zero \cite{lu2023visual}. A gigapixel WSI is transformed into a set of patches (instances), with each patch being embedded into an aligned visual-language latent space. where the similarity scores between the embeddings of patches and the embeddings of prompts are combined using a permutation-invariant operation like topK max-pooling to generate the classification prediction at the WSI level.}
\label{fig:mizero}
\end{figure*}

Visual language pre-training has made significant advancements in representation learning, especially evident in challenging scenarios like zero-shot transfer tasks in open-set image recognition. Nevertheless, computational pathology hasn't delved deeply into zero-shot transfer due to data scarcity and challenges presented by gigapixel histopathology whole-slide images (WSI). Drawing inspiration from the success of multiple-instance learning in weakly supervised learning tasks, \cite{lu2023visual} introduced MI-Zero (\Cref{fig:mizero}). In this method, each WSI is divided into smaller tiles, referred to as instances, which are more manageable for the image encoder. Each instance's cosine similarity scores at the patch level are calculated independently against every text prompt within the latent space. Following this, instance-level scores are combined to generate slide-level scores using a permutation-invariant operator, similar to those in multiple instance learning, such as mean or top K pooling. An optional spatial smoothing step aggregates the information of neighboring patches. When tested on three different real-world cancer subtyping tasks, MI-Zero either matched or outperformed baselines, achieving an average median zero-shot accuracy of $70.2$\%.

In another study, Bannur et al. \cite{bannur2023learning} unveiled the BioViL-T method for biomedical Vision-Language Processing (VLP). Exploiting the data's temporal structure, BioViL-T reached state-of-the-art levels in tasks such as progression classification, phrase grounding, and report generation. Incorporating prior images and reports considerably enhanced the model's efficacy in disease classification and sentence-similarity tasks. The hybrid multi-image encoder in BioViL-T adeptly captured spatiotemporal features, proving valuable for tasks demanding dense visual reasoning over time.

Furthermore, Tiu et al. \cite{tiu2022expert} revealed the potential of self-supervised learning models in pathology detection. Their model, CheXzero, showcased accuracies on par with radiologists in pathology classification. Remarkably, it outdid fully supervised models in detecting certain pathologies and demonstrated adaptability to unannotated pathologies, which weren't specifically included during training. Such results emphasize the strength of contrastive textually prompted models in deciphering medical image interpretation tasks from unannotated data, thus minimizing dependence on extensive labeling.

The body of work presented emphasizes contrastive textually prompted models' indispensable role in medical imaging. They showcase efficiency, performance enhancements, and an uncanny ability to infer intricate medical connotations. These models offer a promising solution to data scarcity, enriching medical image understanding and ultimately optimizing healthcare delivery.

\subsubsection{Generative}
Generative models represent another category within the domain of textually prompted models for medical imaging. These models are designed to generate realistic medical images based on textual prompts or descriptions. They employ techniques such as variational autoencoders (VAEs) and generative adversarial networks (GANs) to understand the underlying distribution of medical images, subsequently creating new samples that correlate with given prompts. These models have shown promise in tasks such as producing images of specific diseases, augmenting training data, and crafting images that adhere to attributes detailed in the prompts. They offer valuable tools for data augmentation, anomaly detection, and creating varied medical image datasets for both training and evaluation. Nonetheless, challenges like capturing the intricacies and variability of medical images, maintaining semantic alignment between generated images and prompts, and addressing ethical concerns tied to fabricated medical images persist and warrant further research.

In a notable study, Yan et al. \cite{yan2022clinical} launched Clinical-BERT, a vision-language pre-training model fine-tuned for the medical sector. Pre-training encompassed domain-specific tasks like Clinical Diagnosis (CD), Masked MeSH Modeling (MMM), and Image-MeSH Matching (IMM). Their research demonstrated that Clinical-BERT outperformed its counterparts, especially in radiograph diagnosis and report generation tasks. Such results emphasize the utility of infusing domain-specific insights during the pre-training phase, thereby refining medical image analysis and clinical decision-making.

Singhal et al. \cite{singhal2023towards} put forth Med-PaLM 2, a state-of-the-art large language model (LLM) targeting expert competence in medical question answering. By blending foundational LLM enhancements with medical-specific fine-tuning and innovative prompting tactics, the team sought to amplify the model's proficiency. Med-PaLM 2 exhibited remarkable progress, registering elevated accuracy and better alignment with clinical utility. When subjected to pairwise ranking assessments, medical practitioners even favored Med-PaLM 2's responses over those of their peers in terms of clinical relevance. This progression signifies the budding potential of LLMs in the realm of medical inquiries, inching closer to rivaling human physicians.

Moor et al. \cite{moor2023med} delved into the creation of Med-Flamingo, a few-shot learner with multimodal capabilities, tailor-made for medical applications. The model underwent pre-training on synchronized and staggered medical image-text data, followed by performance assessment on challenging visual question-answering (VQA) datasets. The outcome revealed that Med-Flamingo augmented generative medical VQA performance by up to 20\%, as per clinician evaluations. Moreover, the model demonstrated prowess in addressing intricate medical queries and furnishing comprehensive justifications, surpassing preceding multimodal medical foundational models. These revelations underscore Med-Flamingo's potential to enrich medical AI paradigms, promote personalized medicine, and bolster clinical decisions.

Collectively, these investigations showcase the strides made in the realm of generative textually prompted models and their implications for the medical sector. Merging domain-specific insights with advancements in language models and multimodal learning techniques has yielded auspicious results in areas like radiograph diagnosis, medical question resolution, and generative medical VQA. Such pioneering works fortify the burgeoning research landscape and chart the course for future innovations in generative models tailored for healthcare applications.

\subsubsection{Hybrid}
Hybrid textually prompted models distinguish themselves through the integration of training paradigms, specifically leveraging both generative and contrastive methodologies.

In a notable study, Chen et al. \cite{chen2023medblip} unveiled a streamlined computer-aided diagnosis (CAD) system tailored for a specific 3D imaging modality, MRI. Drawing inspiration from BLIP-2 \cite{Li2023BLIP2BL}, they crafted a language-image pre-training model that employs bootstrapping to amalgamate 3D medical images with textual data via a query mechanism. At the outset, the researchers deployed a patch embedding that was trainable, bridging the disparity between 3D medical images and a previously trained image encoder. This approach markedly diminished the volume of image data requisite for training. Following this, they introduced the MedQFormer, an innovation that harnesses adjustable queries to align visual attributes seamlessly with the linguistic features demanded by a language model. To round off their methodology, they chose BioMedLM \cite{venigalla2022biomedlm} as the foundational language model and fine-tuned it by harnessing the LoRA technique \cite{hu2021lora}.

An exhaustive suite of experiments, encompassing over 30,000 image volumes sourced from five public Alzheimer’s disease (AD) datasets, affirmed the model's prowess. The results spotlighted its proficiency in zero-shot classification, distinguishing healthy individuals, subjects with mild cognitive impairment (MCI), and those diagnosed with AD. This efficacy underscores the model's potential in executing medical visual question-answering (VQA) tasks with precision.

\subsubsection{Conversational}
Conversational textually prompted models aim to enable interactive dialogues between medical professionals and the model by fine-tuning the foundational models on specific instruction sets. These models facilitate communication and collaboration between humans and the model, allowing medical experts to ask questions, provide instructions, or seek explanations regarding medical images. By incorporating conversational capabilities, these models enhance the interpretability and usability of foundational models in medical imaging. Researchers have explored various techniques to fine-tune the models on conversational datasets and develop architectures that can effectively process textual prompts in a dialogue context. Conversational textually prompted models hold great potential in medical imaging, enabling improved communication, knowledge transfer, and decision-making processes among medical professionals and AI systems. However, challenges related to understanding context, handling ambiguous queries, and ensuring accurate responses in complex medical scenarios are areas that require further investigation and refinement.

In the study conducted by Li et al. \cite{li2023llava}, a cost-efficient approach for training a vision-language conversational assistant for biomedical images was introduced. The researchers leveraged a large-scale biomedical figure-caption dataset and utilized GPT-4 to generate instructions from text alone. By fine-tuning a general-domain vision-language model using a curriculum learning method, they developed the LLaVA-Med model. The findings showed that LLaVA-Med outperformed previous state-of-the-art models on certain metrics in three standard biomedical visual question-answering datasets. This highlights the potential of Conversational Textually Prompted Models, such as LLaVA-Med, in assisting with inquiries and answering open-ended research questions about biomedical images.

Another study, conducted by Thawkar et al. \cite{thawkar2023xraygpt}, focused on the development of XrayGPT, a conversational medical vision-language model designed specifically for analyzing chest radiographs. XrayGPT aligned a medical visual encoder (MedClip) with a fine-tuned large language model (Vicuna) to enable visual conversation abilities grounded in a deep understanding of radiographs and medical knowledge. The study found that XrayGPT demonstrated exceptional visual conversation abilities and a deep understanding of radiographs and medical domain knowledge. Fine-tuning the large language model on medical data and generating high-quality summaries from free-text radiology reports further improved the model's performance. These findings highlight the potential of Conversational Textually Prompted Models like XrayGPT in enhancing the automated analysis of chest radiographs and aiding medical decision-making.

The study by Shu et al. \cite{shu2023medalpaca}, introduced Visual Med-Alpaca, an open-source parameter-efficient biomedical foundation model that combines language and visual capabilities. Visual Med-Alpaca was built upon the LLaMa-7B architecture and incorporated plug-and-play visual modules. The model was trained using a curated instruction set generated collaboratively by GPT-3.5-Turbo and human experts. The findings showed that Visual Med-Alpaca is a parameter-efficient biomedical model capable of performing diverse multimodal biomedical tasks. Incorporating visual modules and using cost-effective techniques like Adapter, Instruct-Tuning, and Prompt Augmentation made the model accessible and effective. This study emphasizes the importance of domain-specific foundation models and demonstrates the potential of conversational textually prompted models like Visual Med-Alpaca in biomedical applications.

\subsection{Visually Prompted Models}
\label{sec:visual}
Within medical imaging, the recent surge of visually prompted models promises a blend of precision, adaptability, and generalization. These models, informed by the extensive capabilities of foundation models, offer the potential to revolutionize medical image analysis by catering to specific tasks while also adapting to a vast array of modalities and challenges. This section delves into two main trajectories of such models:

\begin{enumerate}
    \item \textbf{Adaptations}: As the name suggests, this sub-section explores the adaptations and modifications made to traditional segmentation models, enhancing their specificity and performance for medical imaging tasks. From models that augment SAM's capabilities for medical images to frameworks that synergize few-shot localization with segmentation abilities, we traverse the journey of various innovations in the realm of medical image segmentation.
    
    \item \textbf{Generalist}: Moving beyond task-specific adaptability, the models in this sub-section embody the essence of a 'Generalist' approach. They are designed to encompass a broader spectrum of tasks and data modalities. These models not only process different kinds of medical imaging data but also can integrate patient histories and genomic data, marking a stride towards a more holistic healthcare technology ecosystem.
\end{enumerate}

As we delve deeper into this section, we will uncover the transformative potential of visually prompted models in medical imaging, highlighting both their specialized adaptations and their expansive generalist capabilities.

\begin{figure*}[!thb]
    \centering
    \includegraphics[width=\textwidth]{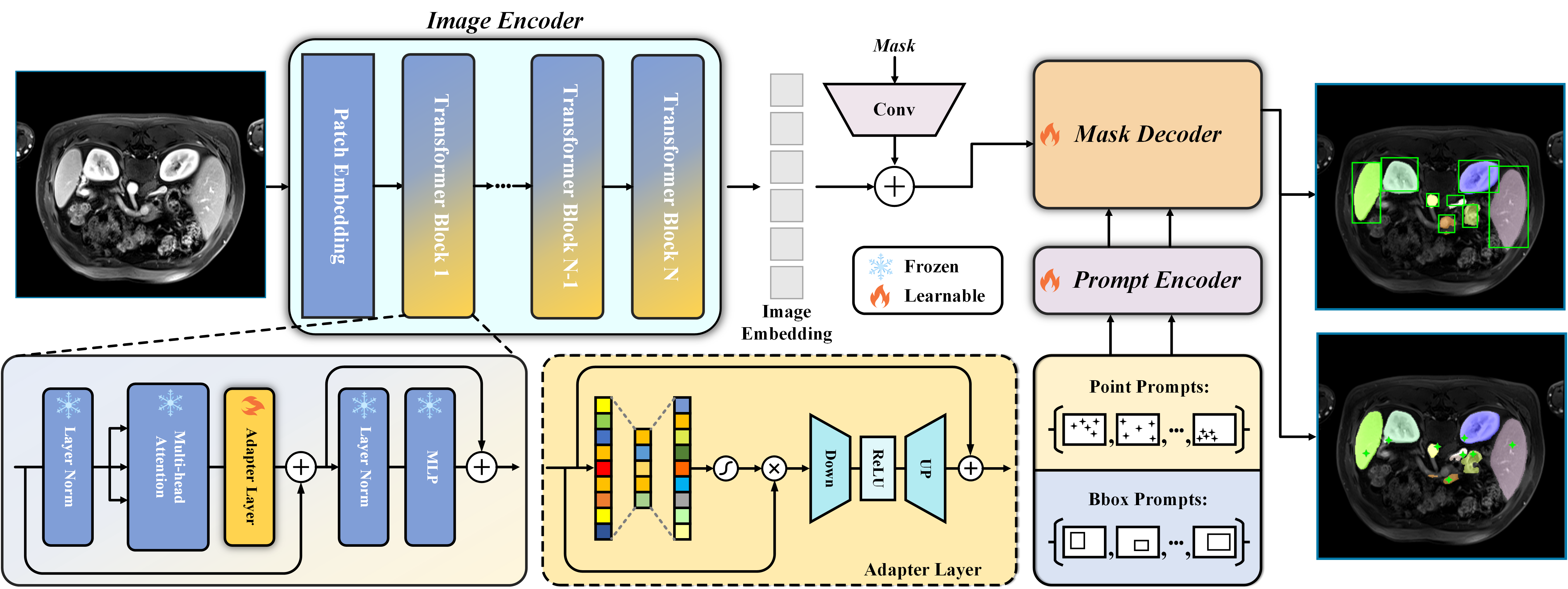}
    \caption{The SAM-Med2D pipeline \cite{cheng2023sammed2d} involves freezing the image encoder and introducing learnable adapter layers within each Transformer block to assimilate domain-specific expertise in the medical domain. The prompt encoder is fine-tuned using point, Bbox, and mask information, with the mask decoder's parameters being updated through interactive training.}
    \label{fig:sam-med2d}
\end{figure*}

\subsubsection{Adaptations}

Traditional medical image segmentation has primarily relied on task-specific models, which, while accurate in their domains, often lack the ability to generalize across multiple tasks and imaging modalities. This necessitates a tailored, resource-intensive approach for each segmentation challenge. The advent of foundation models trained on extensive datasets presents an exciting solution. These models are capable of recognizing and segmenting numerous anatomical structures and pathological lesions across different imaging modalities. However, despite their potential, there are challenges with existing models like SAM, especially when applied to medical images \cite{wu2023medical}. This necessitates further innovations to extend their capabilities, and one such approach is the Medical SAM Adapter, which bridges the gap and enhances SAM's performance in the medical domain \cite{wu2023medical}. This promises an integration of automated processes with specific customization.

Ma and Wang presented MedSAM, a novel foundation model crafted for medical image segmentation \cite{ma2023segment}. Built using a comprehensive dataset of over a million medical image-mask pairs, MedSAM can address numerous segmentation tasks across various imaging modalities. Its promptable configuration seamlessly blends automation with user-driven customization. MedSAM excelled in tasks, especially in computing pivotal biomarkers like accurate tumor volume in oncology. However, it had some limitations, such as modality representation imbalances in its training data and challenges in segmenting vessel-like structures. Nevertheless, its architecture permits future refinements to cater to specific tasks, emphasizing the adaptability of foundation models in medical image segmentation.

Lei et al. tackled the challenge of the intensive annotation workload inherent in the SAM, by introducing MedLSAM, a novel framework that synergizes few-shot landmark localization with SAM's segmentation capabilities \cite{lei2023medlsam}. MedLSAM framework consists of a Localization Anything Model (MedLAM), which employs a shared Pnet to transform support and query patches into 3D latent vectors. During inference, MedLAM initiates with a randomly positioned agent in the query image and guides it toward the target landmark. The agent's trajectory is updated based on a 3D offset computed from the MedLAM model, effectively localizing the landmark coarsely within the query image. This coarse localization is further refined using the Multi-Scale Similarity (MSS) component, enhancing the accuracy of landmark positioning significantly. Having localized the landmarks, the framework transitions to segmentation using both SAM and MedSAM, a specialized version of SAM fine-tuned for medical images. Trained on an extensive dataset of 14,012 CT scans, MedLSAM autonomously generates 2D bounding boxes across slices, facilitating SAM's segmentation tasks. Impressively, it equaled SAM's performance on two 3D datasets that spanned 38 organs but with significantly fewer annotations. Future-proofed with forward compatibility, MedLSAM opens doors for integration with evolving 3D SAM models, signaling even more effective segmentation in the medical domain.

Gong et al. tackled the challenges posed by SAM, originally for 2D natural images when applied to 3D medical image segmentation, particularly for tumor detection \cite{gong20233dsam}. The team introduced a strategy transforming SAM for 3D medical imaging while retaining most of its pre-trained parameters. By employing a visual sampler for the prompt encoder and a lightweight mask decoder emphasizing multi-layer aggregation, the resulting model, the 3DSAM-adapter, exhibited superior performance. It outperformed leading medical segmentation models in three of four tasks, reaffirming the potential to enhance SAM's utility in intricate medical imaging tasks.

\begin{figure*}[!thb]
    \centering
    \includegraphics[width=\textwidth]{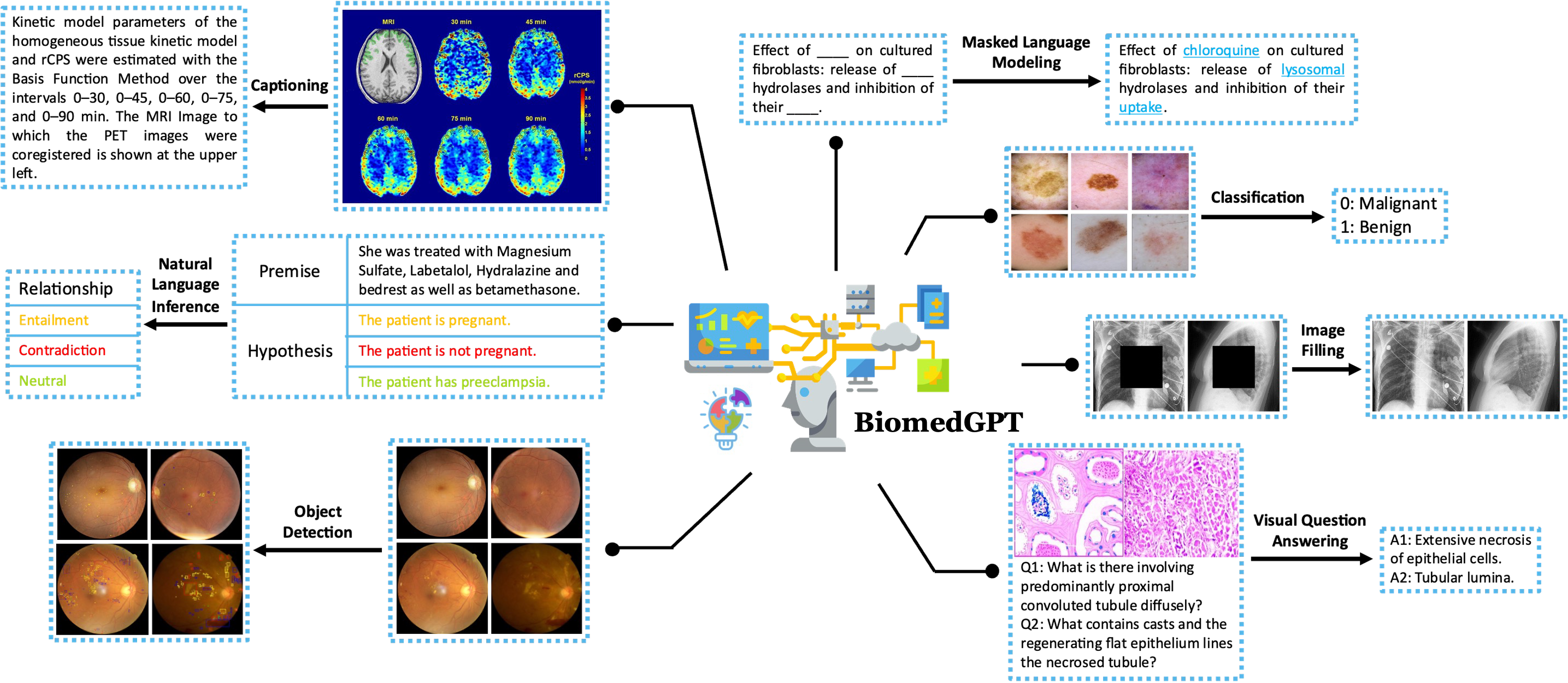}
    \caption{BiomedGPT \cite{zhang2023biomedgpt} demonstrates its versatility in various tasks through pretraining, including unimodal and multimodal approaches, and incorporates object detection for location data. After pretraining, it excels in five downstream tasks, showcasing its data efficiency.}
    \label{fig:biomedgpt}
\end{figure*}

Cheng et al. introduced SAM-Med2D, a specialized model for 2D medical image segmentation \cite{cheng2023sammed2d}. Recognizing the need for domain adaptation, they amassed a substantial dataset of approximately 4.6M images and 19.7M masks, spanning diverse medical modalities. A notable feature of SAM-Med2D is its varied prompt strategies, going beyond bounding boxes and points to incorporate masks, offering a comprehensive interactive segmentation approach as shown in \Cref{fig:sam-med2d}. Thorough evaluations showcased its superior performance across various anatomical structures, with remarkable generalization capabilities proven on datasets from the MICCAI 2023 challenge. Despite its prowess, certain challenges remain, particularly with complex boundaries and low-contrast objects. With prospects of integrating natural language interaction, SAM-Med2D stands as a pioneering contribution to medical computer vision research. Building upon the theme of customization, another noteworthy effort is the development of SAMed. This model, unlike its predecessors, employs a low-rank-based finetuning strategy, enabling it to perform semantic segmentation on medical images with only a fraction of SAM's parameters being updated. This selective approach to parameter adaptation allows SAMed to achieve competitive results, underscoring the potential of customizing large-scale models for specific medical segmentation tasks \cite{zhang2023customized}.

In a stride to enhance reliability in medical image segmentation, Deng et al. put forth SAM-U \cite{deng2023samu}, a novel approach employing multi-box prompts for refined uncertainty estimation in SAM predictions. This method significantly improves SAM's performance, especially in low-quality medical images, and provides crucial insights through generated uncertainty maps, highlighting potential segmentation inaccuracies and serving as an essential guide for clinicians in areas requiring manual annotations. This innovative approach underscores the advancements and adaptability in the realm of medical image segmentation

\subsubsection{Generalist}
In contrast to their adaptability to specific tasks through prompts, foundational models also offer a 'Generalist' approach, further disrupting the landscape of medical imaging. These Generalist models expand upon the foundational model capabilities by being intrinsically designed to handle a broader spectrum of medical imaging tasks and data modalities—ranging from X-rays to MRIs, and even incorporating patient histories and genomic data. The key advantage here is their capability for dynamic task specification, often enabled by natural language descriptions, obviating the need for model retraining. This inherent flexibility is further augmented by the models' ability to formally represent medical knowledge, allowing for reasoned outputs and explanations. The emergence of Generalist models in medical imaging signifies a step towards a more integrated and efficient healthcare technology ecosystem.

Moor et al. \cite{moor2023foundation} delve into the intricacies of developing General-purpose Medical Artificial Intelligence (GMAI), a specialized class of foundation models optimized for the healthcare domain. Unlike conventional medical AI, GMAI models are designed to process multiple data modalities, such as imaging studies and electronic health records, simultaneously. These models are not only capable of complex diagnostic tasks but can also generate treatment recommendations complete with evidence-based justifications. The authors discuss challenges unique to GMAI, including the need for multi-disciplinary panels for output verification and increased susceptibility to social biases due to the complex training data sets. Additionally, they raise concerns over patient privacy and the computational and environmental costs associated with model scaling. The paper underscores that the success of GMAI hinges on rigorous validation and ongoing oversight to mitigate these risks while harnessing its transformative potential in healthcare.

Tu et al. extend the pioneering work of Med-PaLM and Med-PaLM2 \cite{singhal2023towards} to introduce Med-PaLM M, a groundbreaking multi-modal biomedical AI system capable of handling diverse medical modalities, including medical imaging, genomics, and electronic health records \cite{tu2023generalist}. Building upon the foundational achievements of Med-PaLM—which was the first AI to surpass the pass mark on USMLE-style questions—and the subsequent improvements in Med-PaLM2, which boasted an accuracy of 86.5\% on the same questions, Med-PaLM M employs a fusion of Vision Transformer (ViT) for visual tasks and Language-agnostic Language Model (LLM) for natural language tasks. These components are fine-tuned on a newly assembled MultiMedBench dataset. Med-PaLM M eclipses existing benchmarks, including specialized single-task models and its predecessor generalist models like PaLM-E that lacked biomedical fine-tuning. Notably, the system exhibits unprecedented zero-shot learning capabilities, successfully identifying tuberculosis from chest X-ray images without prior training \cite{shi2023generalist}. It also excels in generating radiology reports, rivaling the performance of expert radiologists in human evaluations. While the study highlights the scalability and promise of multi-modal AI models for a range of biomedical tasks, it also acknowledges existing challenges, such as data scarcity and limitations of current benchmarks. The work serves as a seminal contribution, marking a new frontier in biomedical AI, albeit with cautionary notes on safety and equity considerations for real-world applications.

Zhang et al. introduce BiomedGPT \cite{zhang2023biomedgpt}, a unified framework that is trained across multiple modalities—including radiographs, digital images, and text—to perform a diverse range of tasks in the biomedical domain as shown in \Cref{fig:biomedgpt}. The model particularly excels in image classification on MedMNIST v2 datasets and visual question-answering on SLAKE and PathVQA, setting new state-of-the-art benchmarks. However, it lags in text-based tasks such as natural language inference on the MedNLI dataset. One reason for this performance gap is the model's constrained scale; with only 182 million parameters, it is smaller than other state-of-the-art models. The study also pinpoints the model's sensitivity to task instructions and challenges with handling out-of-distribution data as areas for future research. Nonetheless, BiomedGPT represents a significant step towards a versatile, generalist model in the biomedical field, capable of both vision and language tasks.

Wu et al. introduce the Radiology Foundation Model (RadFM) and the MedMD dataset, aiming to unify medical tasks and integrate diverse radiological images \cite{wu2023generalist}. RadFM effectively merges medical scans with natural language, addressing various medical tasks. The study unveils RadBench, a benchmark demonstrating RadFM's superior synthesis of visual and textual information. Despite the advancements, the authors highlight limitations, such as the prevalence of 2D images in the dataset and challenges in generating clinically useful sentences. Wu et al.’s release of these innovations significantly advances radiological models, encourages collaborative progress, and emphasizes the need for enhanced evaluative metrics and comprehensive solutions in the field.

In another study, Zhou et al. introduce RETFound \cite{zhou2023foundation}, a versatile foundation model developed through self-supervised learning, trained on 1.6 million unlabeled retinal images. It demonstrates unparalleled adaptability and generalizability in diagnosing eye diseases and predicting systemic disorders with notable accuracy and reduced reliance on extensive annotation. RETFound overcomes significant barriers related to data limitations and model generalization, offering a pioneering solution in medical AI, with the potential to democratize and significantly advance healthcare AI applications.

\begin{figure*}[!thb]
    \centering
    \includegraphics[width=\textwidth]{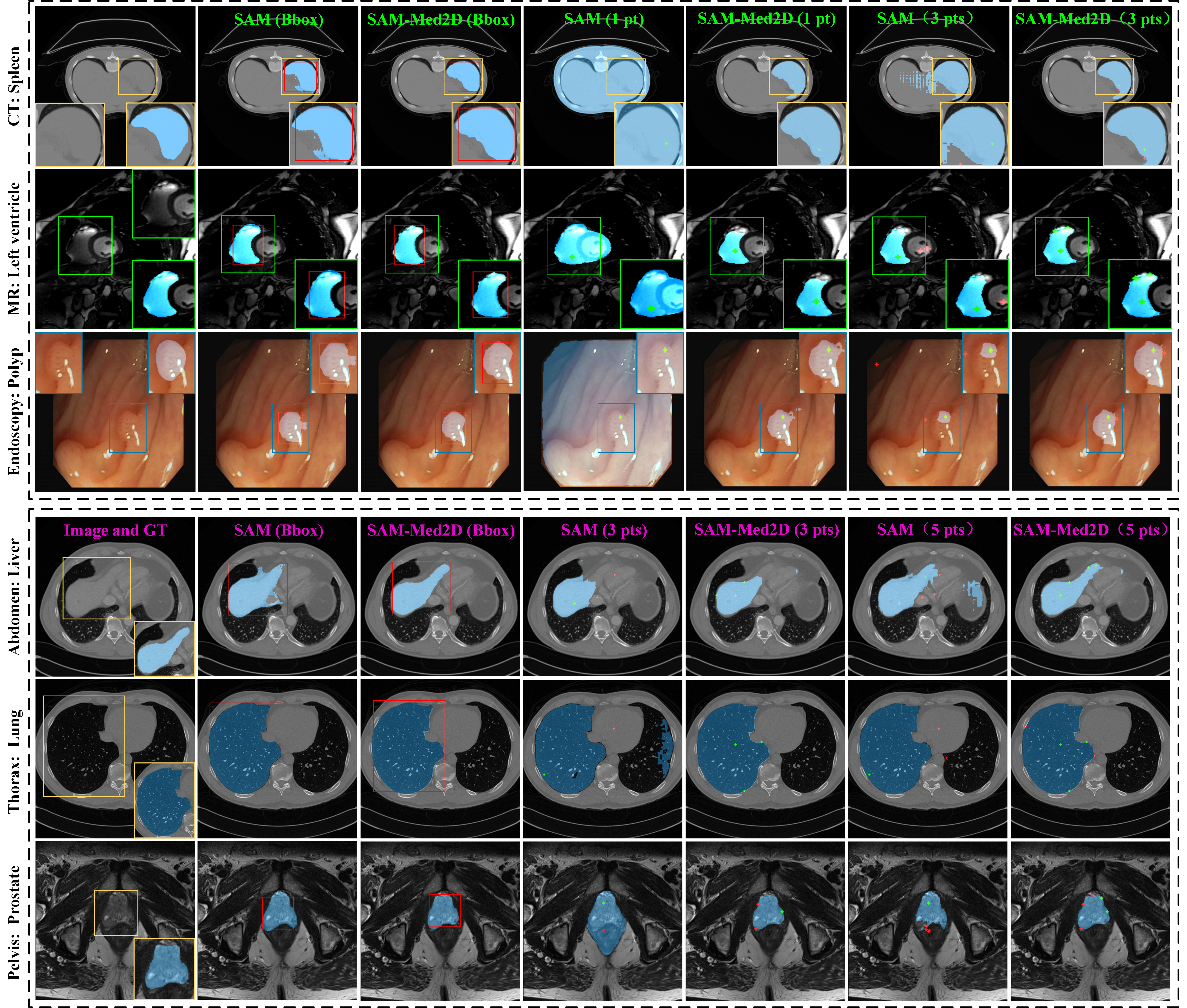}
    \caption{Extensions of the SAM model for diverse medical image segmentation tasks \cite{cheng2023sammed2d}. This figure illustrates the versatility of SAM-based adaptations in addressing a wide range of medical image segmentation challenges, showcasing their applicability and adaptability across various healthcare scenarios.}
    \label{fig:discussion-prove}
\end{figure*}

\section{Discussion}
In the dynamic landscape of foundational models for medical imaging, each direction outlined in our taxonomy (\Cref{fig:taxonomy}) brings its own set of advantages and distinctive capabilities to the forefront. These divergent paths cater to specific needs, creating a diversified toolkit for addressing the multifaceted challenges of the medical imaging domain. As we delve into this discussion, we will explore the unique advantages of each direction and consider scenarios where one direction might excel over the others, all while peering into how these models learn feature representations and the implications thereof.

\noindent \textbf{Textually Prompted Contrastive Models}:
These models have shown remarkable prowess in bridging the semantic gap between medical images and text. By leveraging contrastive learning, these models can extract meaningful representations from unpaired medical image-text data, thereby reducing the dependence on vast amounts of labeled data. This approach is particularly advantageous in scenarios where labeled data is scarce or expensive to obtain, such as rare medical conditions or specialized imaging modalities. Contrastive models excel at capturing subtle medical meanings and are well-suited for tasks like zero-shot prediction in medical image-text tasks. For instance, in scenarios where a new, uncharacterized medical condition arises, these models can adapt swiftly by simply providing textual descriptions.

However, there are limitations to consider. Contrastive models might struggle with highly complex medical images or intricate pathologies, where the nuances demand a deeper level of feature representation. Additionally, they may still rely on the availability of large-scale text data, which could be a bottleneck in some cases. The contrastive learning process also hinges on careful tuning of hyperparameters, making it essential to invest time in fine-tuning for optimal performance.

\noindent \textbf{Textually Prompted Generative Models}:
Textually prompted generative models, exemplified by models like Clinical-BERT \cite{yan2022clinical} and Med-Flamingo \cite{moor2023med}, offer the ability to generate detailed responses and explanations for medical image-related queries. They excel in tasks requiring a deep understanding of the medical domain, making them invaluable in clinical decision support systems, medical education, and generating radiology reports.

These generative models can be a game-changer when interpretability and reasoning are crucial. For instance, in a clinical setting, generating explanations for a model's predictions can enhance trust and facilitate collaboration between AI systems and medical professionals. In educational contexts, they can serve as powerful tutors, providing in-depth explanations and context.

Nevertheless, generative models are computationally intensive and demand significant training data. They may not be the most efficient choice for scenarios where quick, lightweight predictions are required. Additionally, they may face challenges in generating text that is both informative and concise, which could be important in some applications.

\noindent \textbf{Textually Prompted Hybrid Models}:
Hybrid models, as represented by MedBLIP \cite{chen2023medblip}, combine the strengths of generative and contrastive methodologies. These models tackle the challenge of integrating textual data with 3D medical images, often a complex task due to the inherent differences in data modalities.

One of the key advantages of hybrid models is their potential for zero-shot prediction in medical image-text tasks. They seamlessly align visual attributes with linguistic features, making them adept at executing medical visual question-answering tasks with precision. For example, in cases where medical professionals need to quickly diagnose conditions based on both images and textual descriptions, hybrid models can provide valuable support.

Yet, hybrid models may face challenges related to the design of effective integration mechanisms between textual and visual data. The success of these models often relies on the quality of the alignment between different modalities. Additionally, they may require substantial computational resources for training and fine-tuning.

\noindent \textbf{Textually Prompted Conversational Models}:
Conversational models, like LLaVA-Med \cite{li2023llava} and XrayGPT \cite{thawkar2023xraygpt}, are designed to enable interactive dialogues between medical professionals and AI systems. These models are particularly beneficial in scenarios where medical experts need to ask questions, seek explanations, or instruct AI systems regarding medical images.
One of the most significant advantages of conversational models is their potential to enhance communication and collaboration between humans and AI. They can facilitate knowledge transfer, clarify doubts, and provide detailed explanations for complex medical images. In a clinical context, this can lead to more informed decision-making and better patient care.

However, conversational models face the challenge of understanding context and handling ambiguous queries effectively. Ensuring accurate responses in complex medical scenarios remains an ongoing research challenge. Additionally, they require careful fine-tuning on conversational datasets to perform optimally.

\noindent \textbf{Visually Prompted Adaptations Models}:
Visually prompted adaptations models, such as MedLSAM \cite{lei2023medlsam}, MedLSAM, 3DSAM-adapter \cite{gong20233dsam}, SAM-Med2D \cite{cheng2023sammed2d}, SAMed \cite{zhang2023customized}, and SAM-U \cite{deng2023samu}, focus on enhancing the specificity and performance of medical image segmentation tasks. These models adapt foundational models like SAM for the medical domain, addressing challenges like data scarcity and complex boundaries. A sample of segmentation results on various medical image analysis tasks achieved through the adapted SAM model is presented in \Cref{fig:discussion-prove}, showcasing its remarkable achievement in diverse medical imaging scenarios and its robust generalization power.

\begin{table*}[!ht]
	\centering
	\caption{
        Overview of the reviewed Foundation models in medical imaging based on their algorithm choice presented in our taxonomy, \Cref{fig:taxonomy}.
    }
    \label{tab:paperhighlights}
	\resizebox{\textwidth}{!}{
		\begin{tblr}{
				colspec=p{2cm}lp{10cm}p{6cm},
				cells={font=\fontsize{5}{7pt}\selectfont},
				row{1}={font=\bfseries},
				column{1}={font=\bfseries}
		}
			\toprule[2pt]
			\SetRow{purple9} Algorithm & Networks & Core Ideas & Practical Use Cases
            \\ 
            \toprule Textually Prompted Contrastive Models 
            & 
            {
				$^{1}$MedClip \cite{wang2022medclip} \\ 
				$^{2}$BioViL-T \cite{bannur2023learning} \\ 
				$^{3}$CheXzero \cite{tiu2022expert} \\ 
				$^{4}$MI-Zero \cite{lu2023visual}
			} 
            &
			{
                $\bullet$ \cite{wang2022medclip} aims to tackle medical image and report vision-text contrastive learning difficulties. MediCLIP separates medical images and text for multimodal contrastive learning. Scaling combinatorial training data at low cost answers medical data shortages. The study advises replacing InfoNCE with a medical-based semantic matching loss to eliminate contrastive learning false negatives. MedCLIP captures subtle but important medical meanings better than zero-shot prediction, supervised classification, and image-text retrieval methods. The paper reveals MedCLIP's efficacy and data efficiency, which might enhance clinical decision-making and downstream tasks. \\ $\bullet$ Use data temporal structure to improve biological vision-language processing (VLP) in \cite{bannur2023learning}. Researchers introduce BioViL-T, a pre-training system that trains and fine-tunes using past pictures and data. This method uses temporal correlations and a multi-image encoder to handle missing images and longitudinal data without image registration. Modality alignment is improved by analyzing the temporal relationship between visuals and reports, enhancing pre-training and downstream task performance. The study exhibits advanced progression categorization, phrase grounding, and report generation results. Temporal and non-temporal tasks like pneumonia detection and phrase grounding benefit from prior context and temporal knowledge. To test and benchmark chest X-ray VLP models for temporal semantics, the authors offer MS-CXR-T, a multimodal benchmark dataset. An expert radiologist curated this dataset to measure image-text temporal correlations. \\ $\bullet$ In \cite{tiu2022expert}, authors offer a novel medical imaging pathological classifying approach. The study suggests employing self-supervised learning without annotations to accurately diagnose illnesses in unannotated chest X-rays. Large labeled datasets are expensive and time-consuming for traditional medical image interpretation machine-learning algorithms. This research shows that a self-supervised system trained on chest X-rays without annotations can classify illness as well as radiologists. A zero-shot multi-label classification method, natural language supervision from radiology reports, and generalization to diverse image interpretation tasks and datasets are presented in the research. CheXzero learns a representation for zero-shot multi-label classification without labeled data fine-tuning using contrastive learning with image-text pairs. Radiology reports' natural labeling lets self-supervised algorithms perform as well as professional radiologists and fully supervised approaches on unknown disorders. This approach eliminates explicit labeling, eliminating medical machine-learning workflow inefficiencies from large-scale labeling.
            }
			& 
            {
                $\bullet$ Zero-shot prediction in medical image-text tasks, Supervised classification in medical image analysis, Image-text retrieval in the medical domain, Supporting clinical decision-making and downstream clinical tasks \cite{wang2022medclip}.\\ $\bullet$ Progression Classification: Achieving State-of-the-Art Performance in Tracking Medical Condition Progression, Phrase Grounding: Linking Clinical Report Phrases to Image Regions for Enhanced Analysis, Report Generation: Improved Performance by Incorporating Prior Reports, Disease Classification: Consistent Improvement in Disease Classification Tasks, Pneumonia Detection: State-of-the-Art Results in Detecting Pneumonia \cite{bannur2023learning} \\ $\bullet$ Automation of complex medical image interpretation tasks, Disease diagnosis, Diagnostic efficiency improvement, Label efficiency enhancement, Decreased reliance on large labeled datasets, Reduction in labeling efforts and costs, Potential for learning a broad range of medical image interpretation tasks from unlabeled data \cite{tiu2022expert} \\ $\bullet$ Zero-shot transfer for cancer subtype classification on 3 WSI datasets. Moreover, the curated dataset of histopathology image-caption pairs can potentially be generalized and adapted to develop practical solutions in other domains \cite{lu2023visual}.
            } 
            \\ 
            \midrule Textually Prompted Generative Models 
            &
            {
				$^{1}$Clinical-BERT \cite{yan2022clinical} \\ 
				$^{2}$Med-PaLM 2 \cite{singhal2023towards} \\ 
				$^{3}$Med-Flamingo \cite{moor2023med}
			} 
            & 
            {
                $\bullet$ Clinical-BERT \cite{yan2022clinical}, a medical pre-training paradigm, underpins. The research offers domain-specific pre-training activities, including Clinical Diagnosis (CD), Masked MeSH Modeling (MMM), and Image-MeSH Matching for model training. MeSH words in radiograph reports are stressed. The work aligns MeSH terms with radiographs using region and word sparse attention. The model links visual characteristics with MeSH phrases using this attention mechanism. Clinical-BERT radiograph diagnostic and report production provide cutting-edge results. The article shows domain-specific pre-training exercises and MeSH keywords to improve medical task performance. \\ $\bullet$ Expert medical question answering is done using LLMs in \cite{singhal2023towards}. The study aims to enhance LLM performance to match model and clinician replies. The authors say LLMs have advanced in various disciplines and can address medical questions. They admit prior LLM-based models need to be improved, especially compared to clinician responses. The authors offer various LLM performance enhancements. Base LLM improvements (PaLM 2), medical domain-specific fine-tuning, and a new ensemble refinement approach are used. The strategies aim to enhance medical thinking and results. \\ $\bullet$ Med-Flamingo \cite{moor2023med}, a vision-language model suggested is pre-trained on medical image-text data from various sources and can create open-ended replies from textual and visual input. Med-Flamingo outperforms prior models in generative medical visual question-answering tasks by 20\% in clinical assessment scores due to in-context learning The research also describes Visual USMLE, a difficult created VQA dataset including medical questions, images, and case vignettes. The paper says multimodal few-shot and in-context learning improve medical AI models.
            } 
            & 
            {
                $\bullet$ Radiograph Diagnosis and Reports Generation: Achieving state-of-the-art results on challenging datasets, Enhancing Downstream Tasks in the Medical Domain, Improving performance in various medical domain tasks, Learning Medical Domain Knowledge: Enabling the model to acquire domain-specific knowledge for better performance \cite{yan2022clinical} \\ $\bullet$ Medical question answering: Providing accurate and reliable answers to medical questions. Medical exams: Assisting in preparing for medical licensing examinations. Clinical decision support: Aiding physicians in making informed decisions during patient care. Consumer health information: Delivering trustworthy medical information to the general public \cite{singhal2023towards}. \\ $\bullet$ Generative Medical Visual Question Answering (VQA), Medical Reasoning and Rationale Generation, Clinical Evaluation and Human Rater Study, Dataset Creation for Pre-training and Evaluation \cite{moor2023med}
            }
            \\ 
            \midrule Textually Prompted Hybrid Models 
            & 
            {
                $^{1}$MedBLIP \cite{chen2023medblip}
            } 
            & 
            {
                $\bullet$ Extend a 2D image encoder to extract features from 3D medical images and obtain a lightweight language model for our CAD purpose. \\ $\bullet$ Align different types of medical data into the common space of language models, besides collecting the largest public dataset for studying Alzheimer’s disease (AD).
            } 
            &
            {
                $\bullet$ Zero-shot prediction in medical image-text tasks \\ $\bullet$ Zero-shot medical visual question answering (VQA) which involves producing an initial diagnosis for an unseen case by analyzing input images and textual descriptions, while also offering explanations for the decision-making process.
            } 
            \\ 
            \midrule Textually Prompted Conversational Models 
            & 
            {
				$^{1}$LLaVA-Med \cite{li2023llava} \\ 
				$^{2}$XrayGPT \cite{thawkar2023xraygpt} \\ 
				$^{3}$Visual Med-Alpaca \cite{shu2023medalpaca} \\
				$^{4}$PMC-LLaMA\cite{wu2023pmc} \\
				$^{5}$ClinicalGPT \cite{wang2023clinicalgpt} \\
				$^{6}$Radiology-LLamA2 \cite{liu2023radiologyllama2}\\
			} 
            & 
            {
                $\bullet$ Training a low-cost vision-language conversational assistant for biological imagery is the main notion of \cite{li2023llava}. The authors recommend training the computer using a big PubMed Central biomedical figure-caption dataset. Caption data and a novel curriculum learning process let GPT-4 self-instruct open-ended education. The model can align biological vocabulary using figure-caption pairings and grasp open-ended conversational semantics. This strategy resembles how laypeople absorb biological topics. LLaVA-Med can answer biological picture inquiries and has great multimodal communication skills. Fine-tuning LLaVA-Med outperforms supervised biomedical visual question answering in this investigation. The paper releases instruction-following data and the LLaVA-Med model for biomedical multimodal learning research. \\ $\bullet$ XrayGPT \cite{thawkar2023xraygpt}, a conversational medical vision-language model, answers open-ended chest radiograph questions. The model uses MedClip's visual characteristics and Vicuna's textual information to assess radiographs and medical domain knowledge. Interactive and high-quality free-text radiology report summaries enhance XrayGPT automated chest radiograph processing. XrayGPT domain-specific information may enhance chest radiograph analysis. \\ $\bullet$ \cite{shu2023medalpaca} proposes ``visual medical specialists" for multimodal biological activities. Training the model using GPT-3.5 Turbo and human experts use instruction-tuning. Plug-and-play visual modules integrate text and vision for multimodal applications. Visual Med-Alpaca is open-source and cheap for doctors. \\ $\bullet$ \cite{wu2023pmc} uses an open-source medical language model. Through medical expertise, the study proposes a logical strategy to adapt a general-purpose language paradigm to medicine. The language model contains 4.8 million biomedical academic papers and 30,000 medical textbooks. Medical accuracy is improved by fine-tuning the model to domain-specific instructions. Language model reasoning is improved by the paper. The model improves medical judgments by applying medical expertise to case facts and offering well-justified recommendations. Improvement of the language model's alignment ability to adapt to different tasks without task-specific training is also stressed. \\ $\bullet$ To improve NLP, \cite{wang2023clinicalgpt} suggests pre-training and fine-tuning huge language models. Factual errors and a lack of medical language model experience are admitted. A clinical-optimized language model, ClinicalGPT, overcomes these concerns. ClinicalGPT training combines medical records, domain-specific knowledge, and multi-round discussions. This method offers ClinicalGPT context and expertise for clinical tasks. With medical knowledge question-answering, tests, patient consultations, and medical record diagnostic analysis, the study provides a complete evaluation system. This approach assesses ClinicalGPT's medical performance. ClinicalGPT improves with parameter-efficient fine-tuning. For clinical use, these methods improve model parameters. For huge language models in healthcare, ClinicalGPT outperforms others. \\ $\bullet$ \cite{liu2023radiologyllama2} aligns the model with task-specific user objectives, develops radiology-specific language models, evaluates and improves generated impressions and shows the model's better clinical impression-generating performance over other generative language models. The paper says that personalized language models can automate radiology jobs and improve human competency.
            } 
            & 
            {
                $\bullet$ Multimodal Conversational Assistant: LLaVA-Med demonstrates excellent multimodal conversational capability and can assist with inquiries about biomedical images, Biomedical Visual Question Answering (VQA): LLaVA-Med outperforms previous state-of-the-art methods on certain metrics for biomedical VQA tasks, Empowering Biomedical Practitioners: The proposed approach empowers biomedical practitioners by providing assistance with open-ended research questions and improving their understanding of biomedical images \cite{li2023llava} \\ $\bullet$ Automated Analysis: XrayGPT enables automated analysis of chest radiographs, Concise Summaries: XrayGPT provides concise summaries highlighting key findings and overall impressions, Interactive Engagement: Users can engage interactively by asking follow-up questions to XrayGPT, Clinical Decision Support: XrayGPT assists medical professionals in making clinical decisions and provides valuable insights, Advancing Research: XrayGPT opens up new avenues for research in the automated analysis of chest radiographs \cite{thawkar2023xraygpt}. \\ $\bullet$ Interpreting radiological images, Addressing complex clinical inquiries, Providing information on chemicals for hair loss treatment (as a case study), Supporting healthcare professionals in diagnosis, monitoring, and treatment, Enabling prompt generation for specialized tasks (e.g., radiology image captioning) \cite{shu2023medalpaca}
            } 
            \\ 
            \midrule Visually Prompted Adaptations Models 
            &
            {
				$^{1}$MedSAM \cite{ma2023segment} \\ 
				$^{2}$MedLSAM \cite{lei2023medlsam} \\
				$^{3}$3DSAM-adapter \cite{gong20233dsam} \\
				$^{4}$SAM-Med2D \cite{cheng2023sammed2d}\\
				$^{5}$SAMed \cite{zhang2023customized}\\
				$^{6}$SAM-U \cite{deng2023samu}
			} 
            & 
            {
                $\bullet$ Versatile and accurate delineation of anatomical structures and pathologies across various medical imaging modalities, surmounting challenges of modality imbalance and intricate segmentation.\\ $\bullet$ Utilizes merged localization and segmentation with a shared 3D coordinate system for streamlined, precise 3D medical image analysis.\\ $\bullet$ Improves accuracy in decoding spatial patterns in volumetric data through enhanced, lightweight 3D medical image segmentation, focusing particularly on tumors.\\ $\bullet$ Optimized for precise 2D medical image segmentation, utilizing diverse prompts and refinements.\\ $\bullet$ utilizes a low-rank-based finetuning strategy for specialized medical image segmentation, maintaining minimal costs and enhanced capabilities.\\ $\bullet$ Employs multi-box prompts to refine SAM's segmentation with pixel-level uncertainty estimation, increasing accuracy and providing nuanced image understanding.
            } 
            & 
			{
                $\bullet$ Pivotal for a range of clinical applications including efficient segmentation, diagnosis, treatment planning, disease monitoring, and in oncology for accurate tumor volume computation, contributing to personalized patient care and improved health outcomes \cite{ma2023segment}.\\ $\bullet$ A versatile, scalable foundation in medical imaging, reducing annotation burdens, and providing accurate, automated segmentation across medical disciplines, enhancing diagnostic procedures \cite{lei2023medlsam}.\\ $\bullet$ Facilitates clinical diagnosis, treatment planning, and medical R\&D through improved segmentation and serves as a blueprint for domain-specific adaptations, enhancing medical imaging automation and processes \cite{gong20233dsam}.\\ $\bullet$ Enables accurate medical image analysis, offering insights for researchers and advancing medical computer vision and interactive segmentation \cite{cheng2023sammed2d}.\\ $\bullet$ Serves as a crucial tool in computer-assisted medical diagnoses, excelling in multi-organ segmentation tasks, and is fully compatible with the existing SAM system, offering enhanced accessibility and utility in real-world medical settings \cite{zhang2023customized}.\\ $\bullet$ Valuable for providing pixel-level uncertainty estimation in segmentation, aiding precise diagnoses, and identification of segmentation errors, especially in fundus images. It enriches clinical analyses and fosters the development of advanced segmentation methods \cite{deng2023samu}.
            } 
            \\ 
            \midrule Visually Prompted Generalist Models 
            & 
            {
				$^{1}$GMAI \cite{moor2023foundation} \\
				$^{2}$BiomedGPT \cite{zhang2023biomedgpt} \\
				$^{3}$Med-PalM M \cite{tu2023generalist} \\
				$^{4}$RadFM \cite{wu2023generalist} \\
				$^{5}$RETFound \cite{zhou2023foundation}
			} 
            & 
            {
				$\bullet$ Utilizes self-supervision on diverse datasets for multifunctional medical tasks with minimal labeled data, adapting to new tasks and enabling dynamic interaction and advanced reasoning.\\
				$\bullet$ excels in diverse tasks with one model weight set, surpassing specialized models and offering versatile zero-shot generalization in biomedicine.\\
				$\bullet$ Utilizes multi-task pretraining for knowledge transfer to unseen data, aiming to establish new benchmarks in biomedicine.\\
				$\bullet$ Adeptly integrates and analyzes multidimensional medical scans with natural language.\\
				$\bullet$ Uses masked autoencoder techniques to identify retinal structures and patterns related to eye and systemic diseases like heart failure, showing high adaptability across various tasks.
			} 
            & 
			{
				$\bullet$ Has potential applications in generating radiology reports and aiding medical procedures, reducing radiologist workload through automated, contextual report drafting and visualization, and supporting surgical teams with real-time annotations, alerts, and medical reasoning, thereby improving healthcare delivery \cite{moor2023foundation}.\\
				$\bullet$ Efficiently conducts tasks such as image classification and report generation, supporting clinical decisions and diagnostics. It serves versatile medical needs, offering reliable interpretations of diverse biomedical data, particularly where specialized models are unattainable and integrated insights are vital \cite{zhang2023biomedgpt}. \\
				$\bullet$ Uncovers insights essential for healthcare advancements by integrating information from various medical fields for diverse applications and analyses, requiring no finetuning modifications, and efficiently solving real-world problems \cite{tu2023generalist}. \\
				$\bullet$ It integrates multiple images, essential for longitudinal follow-ups and diverse scenarios, aiding professionals in generating accurate, context-rich reports and plans by understanding both visual and textual medical data \cite{wu2023generalist}. \\
				$\bullet$ Useful in clinical settings for early detection and risk assessment, it offers a data-efficient solution, minimizing annotation efforts and promoting broader implementation in varied clinical applications, contributing to the democratization of advanced healthcare AI technologies \cite{zhou2023foundation}.
			} 
            \\ 
            \bottomrule[2pt]
		\end{tblr}
	}
\end{table*}

The primary advantage of adaptation models is their ability to excel in specialized medical image segmentation tasks. For instance, in scenarios where precise tumor volume calculation is critical, models like MedSAM can provide accurate results. These models are tailored for the medical domain, making them well-suited for specific clinical applications.

Nonetheless, adaptation models may require substantial annotated data for fine-tuning. They might face challenges in scenarios with limited labeled data, as achieving the desired level of performance could be challenging. Additionally, they might not be the most efficient choice for tasks that require generalization across diverse medical imaging modalities.

\noindent \textbf{Visually Prompted Generalist Models}:
Visually prompted generalist models, exemplified by models like BiomedGPT \cite{zhang2023biomedgpt} and Med-PalM M \cite{tu2023generalist}, offer versatility by handling a wide spectrum of medical imaging tasks and data modalities. They can seamlessly switch between tasks without the need for extensive retraining, making them suitable for dynamic healthcare environments.

The key advantage of generalist models is their flexibility. In scenarios where medical professionals need a single model that can handle various tasks, such as image classification, text generation, and question-answering, these models shine. Their ability to reason across different modalities and provide informed responses is invaluable in clinical decision support and medical research.

However, generalist models might face challenges related to task-specific fine-tuning. Achieving state-of-the-art performance in highly specialized tasks might require additional domain-specific data. Moreover, these models need robust mechanisms for handling out-of-distribution data effectively.

\begin{figure*}[!thb]
\centering
\includegraphics[width=0.8\textwidth]{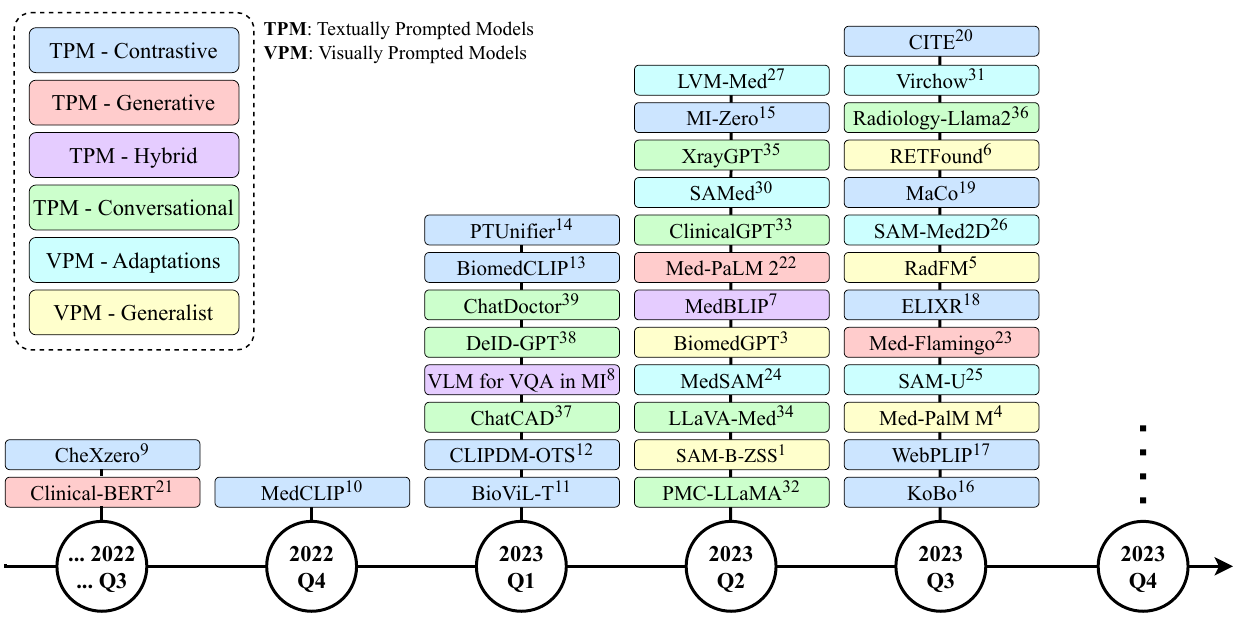}
\caption{Timeline of advancements in textually and visually prompted foundation models in medical imaging over the past quarters." This timeline illustrates significant progress and breakthroughs in the field, spanning the last five quarters, highlighting the dynamic nature of research and innovation in textually and visually prompted foundation models for medical image analysis.}
\label{fig:timline}
\end{figure*}

In conclusion, the choice between these directions largely depends on the specific use case and requirements. While textually prompted models excel in tasks requiring interpretation and detailed explanations, visually prompted models dominate in segmentation and image-specific tasks. Conversational models bridge the gap between human experts and AI systems, facilitating collaborative decision-making. The choice ultimately boils down to the nature of the problem, the availability of data, and the need for adaptability or versatility in the medical imaging domain. Each direction contributes to the evolving landscape of foundational models, offering a rich tapestry of tools to tackle diverse healthcare challenges. 
To facilitate comprehension, we have presented the benefits, drawbacks, and practical applications of each direction in \Cref{tab:paperhighlights}. 
We also showcase the timeline of the reviewed papers in the past quarters, as illustrated in \Cref{fig:timline}.
This figure presents a chronological overview of the key milestones and developments in the field, highlighting the rapid evolution and growing significance of textually prompted and visually prompted foundation models in medical imaging. Expanding on this, the timeline provides valuable insights into the progression of research, revealing the emergence of novel techniques, conversational, and generalist models.

\subsection{Hardware Requirements and Dataset Sizes}
In the pursuit of advancing foundational models in medical imaging, it is imperative to consider the practical aspects of implementing these models in real-world healthcare settings. Two crucial aspects in this regard are the hardware requirements and dataset sizes, both of which significantly influence the feasibility and scalability of deploying these models.
\\
\noindent\textbf{Hardware Requirements}:
Many foundational models, owing to their immense complexity and scale, have substantial hardware requirements. While these models deliver remarkable performance, their training and inference often demand significant computational resources. For instance, models like BiomedGPT, Clinical-BERT, and Visual Med-Alpaca, with millions to billions of parameters, necessitate high-end GPUs or even dedicated hardware accelerators for efficient operation. It is essential to acknowledge that the hardware investment required for these models may present a challenge for resource-constrained healthcare institutions. Therefore, striking a balance between model performance and hardware feasibility is a crucial consideration when implementing these models in clinical practice. Future research should explore strategies to optimize these models for deployment on less resource-intensive hardware, making them accessible to a wider range of healthcare facilities.
\\
\noindent\textbf{Dataset Sizes}:
Another noteworthy aspect is the size of the datasets used to train and fine-tune foundational models. Larger datasets often result in improved model performance and generalization, but they can be challenging to obtain in the medical domain due to privacy concerns and the labor-intensive nature of medical data annotation. Several papers in our survey have employed datasets with varying sizes, from thousands to millions of medical images and reports. Understanding the dataset size requirements for achieving state-of-the-art results is vital for healthcare practitioners and researchers. While some models demonstrate exceptional performance with relatively small datasets, others rely on extensive datasets to excel in complex medical tasks. Future research should explore techniques for efficient dataset collection, augmentation, and utilization, enabling the development of models that can perform well with limited data while preserving patient privacy.

To provide a more detailed overview of the hardware requirements and dataset sizes reported in the reviewed papers, we present \Cref{tab:gpu-demands}. In this table, we present sample hardware configurations utilized for training the network, alongside details regarding dataset sizes.

\begin{table*}
    \centering
    \caption{A summary of publicly available information about medical foundational models,
their computational demands and training information. The unavailable information is featured with a dash.}
    \label{tab:gpu-demands}
    \resizebox{\textwidth}{!}{
    \begin{tblr}{
      width = \textwidth,
      colspec = {llllllllllll},
      cells = {lskyblue},
      row{1} = {lgray},
      row{10-11} = {llightred},
      row{12-13} = {PinkLace},
      row{14-18} = {llightgreen},
      row{19-24} = {llightcyan},
      row{24-26} = {llyellow},
      hlines,
      vline{1,13} = {-}{},
    }
    \textbf{ID} & \textbf{Category} & \textbf{Sub-category} & \textbf{Short name} & \textbf{GPU Model} & {\textbf{Number of}\\\textbf{~GPUs}} & {\textbf{GPU Memory}\\\textbf{~(GB)}} & {\textbf{Total GPU }\\\textbf{Memory (GB)}} & {\textbf{Training Time }\\\textbf{(GPU Hour)}} & \textbf{Input Size} & {\textbf{Total }\\\textbf{batch size}} & \textbf{Epochs}\\
    \textbf{1} & \textbf{TPM} & Contrastive & MedCLIP & Nvidia RTX 3090 & 1 & 24 & 24 & 8 & 224x224 & 100 & 10\\
    \textbf{2} & \textbf{TPM} & Contrastive & BioViL-T & Nvidia Tesla V100 & 8 & 32 & 256 & - & 448x448 & 240 & 50, 100 \\
    \textbf{3} & \textbf{TPM} & Contrastive & CLIPDM-OTS & NVIDIA RTX A5000 & 8 & 24 & 192 & - & 96x96x96 & 42 & 50 \\
    \textbf{4} & \textbf{TPM} & Contrastive & PTUnifier & NVIDIA A100 & 4 & 80 & 320 & - & 288×288-384x384 & 16-128 & 11-60 \\
    \textbf{5} & \textbf{TPM} & Contrastive & BiomedCLIP & NVIDIA A100 & 16 & 40 & 640 & - & 224x224-336x336 & 4k-64k (context) & 40 \\
    \textbf{6} & \textbf{TPM} & Contrastive & KoBo & Nvidia RTX 3090 & 2 & 24 & 48 & - & - & 100 & 50 \\
    \textbf{7} & \textbf{TPM} & Contrastive & MI-Zero & NVIDIA A100 & 8 & 80 & 640 & - & 448x448 & 512 & 50 \\
    \textbf{8} & \textbf{TPM} & Contrastive & CITE & GeForce GTX 2080 Ti  & 2 & 11 & 22 & 0.37 & 224x224 & 128 & (1000 iteration) \\
    \textbf{9} & \textbf{TPM} & Generative & Clinical-BERT & Nvidia RTX 3090 & 2 & 24 & 48 & 96 & 224x224 & 256 & 50 \\
    \textbf{10} & \textbf{TPM} & Generative & Med-Flamingo & Nvidia A100 & 8 & 80 & 640 & 1296 & - & 400 & - \\
    \textbf{11} & \textbf{TPM} & Hybrid & MedBLIP & Nvidia RTX 3090 & 1 & 24 & 24 & - & 224x224x224 & 7 & 100 \\
    \textbf{12} & \textbf{TPM} & Hybrid & VLM for VQA in MI & GeForce GTX 1080 Ti  & 1 & 11 & 11 & - & 224x224 & 50 & 50 \\
    \textbf{13} & \textbf{TPM} & Conversational & DeID-GPT & Nvidia RTX 3090 & $>$1 & 24 &  &  &  &  & \\
    \textbf{14} & \textbf{TPM} & Conversational & ChatDoctor & Nvidia A100 & 6 & 80 & 480 & 18 & max-sq-len: 2048 & 192 & 3 \\
    \textbf{15} & \textbf{TPM} & Conversational & PMC-LLaMA & Nvidia A100 & 32 & 80 & 2560 & - & max-sq-len: 2048 & img:256, text:3200 & 8 \\
    \textbf{16} & \textbf{TPM} & Conversational & LLaVA-Med & Nvidia A100 & 8 & 40 & 320 & 120 & - & 128 & ~100 \\
    \textbf{17} & \textbf{TPM} & Conversational & Radiology-Llama2 & Nvidia A100 & 4 & 80 & 320 & - & - & 128 & - \\
    \textbf{18} & \textbf{VPM} & Adaptations & SAMed & Nvidia RTX 3090 & 2 & 24 & 48 & - & 512x512 & 12 & 200 \\
    \textbf{19} & \textbf{VPM} & Adaptations & MedSAM & Nvidia A100 & 20 & 80 & 1600 & - & 1024x2014 & 160 & 100 \\
    \textbf{20} & \textbf{VPM} & Adaptations & AutoSAM & NVIDIA Tesla V100 & 1 & 16 & 16 & - & 1024x1024 & 4 & 120 \\
    \textbf{21} & \textbf{VPM} & Adaptations & LVM-Med & Nvidia A100 & 16 & 80 & 1280 & 2688 & 224x224 - 1024x1024 & 16 - 64 & 20-200 \\
    \textbf{22} & \textbf{VPM} & Adaptations & SAM-Med2D & Nvidia A100 & 8 & 80 & 640 & - & 256x256 & - & 12 \\
    \textbf{23} & \textbf{VPM} & Generalist & SAM-B-ZSS & Nvidia RTX 3080 & 1 & 10 & 10 & - & 1024x1024 & 1 & 20 \\
    \textbf{24} & \textbf{VPM} & Generalist & RadFM & Nvidia A100 & 32 & 80 & 2560 & - & 256(3D), 512(2D) & 1(3D), 4(2D) & 8 \\
    \textbf{25} & \textbf{VPM} & Generalist & RETFound & Nvidia A100 & 8 & 40 & 320 & 2688 & 16x16 & 16, 1792 & 50, 800
\end{tblr}
}
\end{table*}

\section{Open challenges and Future Direction}
\label{sec:challenge}
Throughout this survey, we have conducted an in-depth analysis of various foundational models, delving into their architectural designs, motivations, objectives, and use cases, all aimed at tackling real-world challenges. In this section, our focus shifts to underscore research directions that have the potential to further empower these models for addressing medical imaging applications.


\subsection{Open-source Multimodal Models}
The future direction of foundation models in medical imaging holds immense promise, primarily due to their seamless integration of diverse data modalities. This integration creates opportunities to explore medical concepts at multiple scales and leverage insights from various knowledge sources, including imaging, textual, and audio data. This multimodal integration empowers medical discoveries that are challenging to achieve with single-modality data alone, while also facilitating knowledge transfer across domains \cite{moor2023foundation}. For example, current self-supervised learning methods are not universally generalizable and often need to be tailored and developed for each specific modality, highlighting the ongoing need for research and innovation in this area. Foundation models are poised to revolutionize healthcare by offering a holistic understanding of diseases and enabling more precise and data-driven medical interventions.
However, to truly unlock the full potential of foundation models in this context, we must emphasize the need to consider inter-modality and cross-modality relationships more effectively. This involves developing methods that can effectively bridge the gap between different data modalities, allowing for better information fusion and more accurate predictions. By enhancing the ability to capture the intricate connections between different medical data, we can further increase the performance and utility of foundation models in medical imaging and healthcare. This interdisciplinary approach is critical for advancing our understanding of complex diseases and improving patient care.

\subsection{Interpretablity}
Understanding a model's capabilities, reasoning, and mechanisms provides profound insights into its outputs. Explainability and interpretability are pivotal in adopting foundation models for building trustworthy AI-driven systems and ensuring their ethical and practical use in healthcare \cite{azad2022medical}. These capabilities are essential for transparency, accountability, and regulatory compliance. Specifically, understanding what a model can do, why it behaves in certain ways, and how it operates is particularly vital when dealing with foundation models. These complex models, powered by extensive data, possess the ability to perform unforeseen tasks in entirely novel ways \cite{bommasani2021opportunities}.
In healthcare, explainability is critical for decisions regarding patient symptoms, clinical trials, and informed consent. Transparent AI reasoning helps resolve disagreements between AI systems and human experts, explaining the reasons behind the created decisions. However, most current foundation models lack built-in explainability, requiring future research. By connecting AI outputs with medical knowledge, models become more understandable, enabling users to grasp not only what the model predicts but why. This interdisciplinary approach, merging AI with domain expertise, advances disease understanding, elevates patient care, and promotes responsible AI use in healthcare.

\subsection{Bias and Variance in Foundational Models}
Within the domain of foundational models for medical imaging, two critical aspects demand ongoing scrutiny and investigation: bias and variance \cite{yang2020rethinking}.

\noindent\textbf{Bias}: One of the foremost challenges facing foundational models is the presence of bias in both data and predictions. Just as in vision and language models, foundational models in medical imaging can inherit and amplify biases present in the training data. These biases might be related to race, ethnicity, gender, or socioeconomic factors, and they can manifest in the models' predictions and behaviors. For instance, a model might exhibit disparities in disease diagnosis or treatment recommendations for different demographic groups, potentially leading to unequal healthcare outcomes. Thus, addressing and mitigating biases in foundational models is of paramount importance to ensure fairness, inclusivity, and ethical deployment in the medical domain.

\noindent\textbf{Variance}: Variance, on the other hand, pertains to the models' sensitivity to fluctuations in the training data. In the context of medical imaging, variance can manifest as the models' inability to generalize effectively across diverse patient populations or different healthcare settings. Models with high variance might perform exceptionally well on one dataset but poorly on another, hindering their reliability in real-world clinical applications. Therefore, strategies that enhance the robustness and generalization capabilities of foundational models are crucial for their widespread adoption and utility.

\subsection{Adversarial Attacks}
In the healthcare system, where the accuracy of medical decisions can have life-altering consequences, susceptibility to adversarial attacks \cite{maus2023black} is a pressing concern of paramount importance. These attacks, which involve the deliberate manipulation of model inputs, can lead to not only erroneous but potentially harmful outputs, creating a perilous landscape for medical practitioners and patients alike. For instance, in the context of medical imaging, adversarial attacks could potentially result in misdiagnoses, causing patients to receive incorrect treatments or delay necessary interventions. Furthermore, the compromise of patient data privacy through adversarial tactics can lead to severe breaches of confidentiality, raising ethical and legal concerns.

Additionally, the potential for the spread of false medical information, fueled by adversarial attacks, could have far-reaching consequences, undermining public trust in foundational models and healthcare systems. Therefore, addressing these vulnerabilities and developing robust defence mechanisms are not just academic endeavors but essential imperatives for ensuring the safety, reliability, and ethical use of foundational models in medical applications. The healthcare domain demands a proactive stance in fortifying foundational models against adversarial threats to safeguard the integrity and efficacy of clinical decision-making processes and the privacy of patient data.

\subsection{Down-stream Task Adaptation}
Foundation models offer powerful adaptability, including fine-tuning and prompting, making them versatile for healthcare and medical tasks. However, their extensive initial training demands substantial resources, adapting them efficiently for different tasks without losing learned knowledge is a critical challenge, and there is a need for research to reduce computational and memory requirements for quick adaptation, as current approaches often require careful hyperparameter selection that can impact generalization performance. Therefore, these challenges point to the need for more efficient foundation models in the future to enhance their general-purpose utility.

\subsection{Extensive Data and Computational Demands}
Foundation models, while powerful, come with substantial computational costs for development, training, and deployment. In specific cases, smaller models can achieve similar or better results at a lower cost. Training large-scale models is data and compute-intensive, and acquiring extensive labeled data can be expensive and time-consuming, especially for specialized domains or less-resourced languages. Inference with these models is also costly due to their many parameters. A summary of the computational budget and training costs of some of the reviewed models in this paper is provided in \Cref{tab:gpu-demands}.

These computational demands hinder their practicality in real-world applications, particularly those needing real-time inference or running on resource-constrained edge and mobile devices. For instance, visual prompt-based models like Segment Anything \cite{kirillov2023segment}, while having robust image encoders, currently lack real-time processing speed, a crucial requirement for practical use. FastSAM \cite{zhao2023fast}, on the other hand, achieves comparable performance to the SAM method but at 50 times faster run-time speed by replacing the Transformer architecture with YOLOv8-seg \cite{JocherYOLObyUltralytics2023}, significantly expanding the utility of such models in real-world scenarios. Consequently, there's potential to develop more efficient successors to address this issue, particularly in medical applications where running models on edge devices offer substantial advantages, especially in underserved areas.

\subsection{Prompt Engineering}
Prompt engineering is a critical aspect of foundational models in medical imaging, and its significance lies in its potential to bridge the gap between these models and radiologists, ultimately enhancing patient care \cite{Liu2023DeIDGPTZM}. In the context of medical image interpretation, effective communication between radiologists and AI models can lead to several noteworthy benefits.
First and foremost, prompt engineering allows radiologists to have natural and interactive conversations with AI models. This capability is particularly valuable as it enables radiologists to seek clarifications, provide additional context, and ask follow-up questions, mirroring real-world clinical scenarios. For example, when reviewing a complex medical image, a radiologist may need to ask the AI model for further explanations about its findings, request alternative views, or explore differential diagnoses. Prompt engineering facilitates this conversational flow, making AI models more accessible and collaborative tools for radiologists.
Moreover, the ability to converse with AI models through well-constructed prompts empowers radiologists with a more interactive and intuitive workflow. Instead of relying solely on fixed queries or predefined prompts, radiologists can tailor their interactions based on the specific nuances of each case. This adaptability allows for a more dynamic and personalized user experience, ultimately improving diagnostic accuracy and efficiency.
Furthermore, prompt engineering contributes to the interpretability and transparency of AI models. Radiologists can gain insights into how the model arrives at its conclusions by crafting prompts that elicit detailed explanations. This transparency is crucial in a clinical context, where radiologists need to understand the reasoning behind the AI model's recommendations and trust its diagnostic insights.

\subsection{Lack of Effective Benchmark to Monitor Progress}
While various benchmark datasets and evaluation metrics exist, they often fall short in comprehensively assessing model performance across diverse medical imaging tasks, modalities, and real-world clinical scenarios. Addressing this issue and establishing a robust benchmarking framework is crucial for several reasons.
Firstly, a comprehensive benchmark can facilitate fair and standardized model evaluation, enabling researchers to assess the true strengths and weaknesses of different foundational models accurately. Currently, models may excel in specific datasets or tasks but struggle when applied to new, untested scenarios. An effective benchmark should encompass a wide spectrum of medical imaging challenges, including rare conditions and diverse patient populations, to provide a holistic assessment of model capabilities.
Secondly, a well-structured benchmark can drive innovation by defining clear objectives and goals for the field. It can serve as a reference point for researchers and encourage the development of models that can address real-world clinical needs effectively. Moreover, it can incentivize the creation of models that are robust, interpretable, and adaptable to the dynamic nature of healthcare.
Lastly, an effective benchmarking framework can aid in the deployment of foundational models in clinical practice. By thoroughly evaluating models' performance and generalization across various clinical settings, it can assist healthcare providers in selecting the most suitable models for specific tasks and ensure that AI-assisted medical decision-making is reliable and safe.

\subsection{Enhancing Feature Representation from Frequency Perspective}
Given that the majority of foundational models employ ViT models as their backbone, it becomes crucial to assess these models from a frequency perspective to ensure their ability to capture and learn diverse frequency information necessary for object recognition.
Recent research has shed light on the fact that traditional self-attention mechanisms in ViT, while effective in mitigating local feature disparities, tend to neglect vital high-frequency details, such as textures and edge characteristics \cite{wang2022antioversmoothing,azad2023laplacian}. This oversight is particularly problematic in tasks like tumor detection, cancer-type identification through radiomics analysis, and treatment response assessment, as these tasks often hinge on recognizing subtle textural abnormalities.
Additionally, it's worth noting that self-attention mechanisms come with a quadratic computational complexity and may generate redundant features \cite{azad2023beyond}. Given these considerations, the design of new foundational models should take these limitations into account and explore potential enhancements. This could involve incorporating CNN layers or adopting more efficient ViT architectures to strike a balance between computational efficiency and preserving high-frequency information.
\section{Conclusion}
\label{sec:conclusion}
In this comprehensive survey, we have conducted an in-depth review of recent advancements in foundational models for medical imaging. Our survey commences with an introductory section that provides insight into the evolution of foundation models and their potential contributions to the healthcare sector.

Subsequently, we categorize these models into four main groups, differentiating between those prompted by text and those guided by visual cues. Each of these categories boasts unique strengths and capabilities, and in \Cref{chapter3}, we delve into these directions by presenting exemplary works and offering comprehensive methodological descriptions.
Furthermore, our exploration extends to evaluating the advantages and limitations inherent to each model type. We shed light on their areas of excellence and identify areas where they have room for improvement. This information is presented in the form of pros, cons, and real-world use cases of these models in the context of medical imaging scenarios, and the summarized results can be found in \Cref{tab:paperhighlights}.
Moreover, we consider the hardware and dataset requirements for implementing these models. We provide various configuration strategies to elucidate the prerequisites for future research endeavors, helping researchers gain a clear understanding of the necessary resources.
In conclusion, our survey not only reviews recent developments but also sets the stage for future research in foundational models. We propose several directions for future investigations, offering a roadmap for researchers to excel in the field of foundational models for medical imaging.

\bibliographystyle{unsrt}
\bibliography{Ref}

\end{document}